\RequirePackage{fix-cm}

\documentclass{article} 
\usepackage[OT1]{fontenc}
\PassOptionsToPackage{table}{xcolor}
\usepackage{iclr2027_conference,times}
\usepackage{booktabs}
\usepackage{diagbox}
\usepackage{multirow}
\usepackage{xcolor}
\usepackage{graphicx}
\usepackage{enumitem}
\definecolor{groupgreen}{RGB}{226,240,226}
\usepackage{amsmath}
\usepackage{amssymb}

\usepackage{amsmath,amsfonts,bm}

\def\eqref#1{equation~\ref{#1}}

\def\1{\bm{1}}

\DeclareMathAlphabet{\mathsfit}{\encodingdefault}{\sfdefault}{m}{sl}
\SetMathAlphabet{\mathsfit}{bold}{\encodingdefault}{\sfdefault}{bx}{n}

\usepackage{float}
\usepackage{hyperref}
\usepackage{url}
\usepackage{graphicx}
\usepackage{booktabs}
\usepackage{float}
\usepackage{placeins}

\title{MOBA-VL: Event-Localized\\
Multi-Turn Reinforcement Learning for\\
Real-Time MOBA Commentary}

\author{%
Shengyun Zhong$^{1,2*\dagger}$ \quad Xinkang Zhao$^{1*}$ \quad
Ziyuan Chu$^{1}$ \quad Linchao Zhu$^{1\ddagger}$\\
\normalfont $^{1}$Zhejiang University, China \qquad
$^{2}$Northeastern University, USA
}
\usepackage{fvextra}
\usepackage{CJKutf8} 
\iclrfinalcopy 
\begin{document}
\raggedbottom

\maketitle
\lhead{Preprint}
\begingroup
\renewcommand{\thefootnote}{\fnsymbol{footnote}}
\footnotetext[1]{Equal contribution.\quad $^{\dagger}$Work done during an internship at Zhejiang University.\quad $^{\ddagger}$Corresponding author.}
\endgroup

\vspace{-16pt}
\begin{figure}[H]
    \centering
    \includegraphics[width=0.90\textwidth]{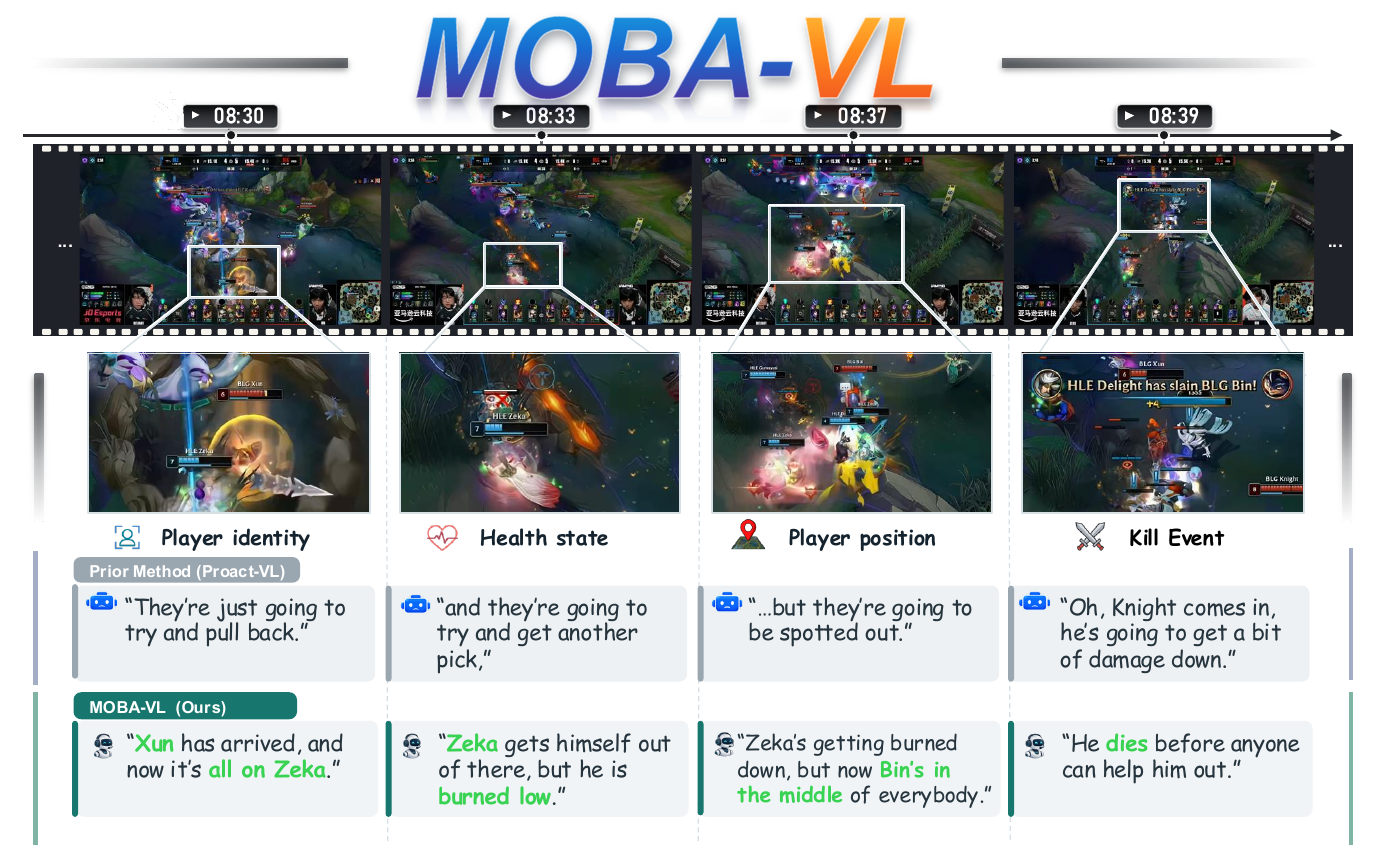}
    \caption{\textbf{Real-time commentary by MOBA-VL.}
Given a 1~FPS video stream, the model generates one commentary
turn per second using only frames observed so far. Four selected
turns from a ten-second sequence are shown. Gray boxes highlight
relevant visual evidence and green text marks player references
and actions. MOBA-VL describes the elimination at 08:39, while
the prior method does not mention it in the displayed turn.}
    \label{fig:teaser}
\end{figure}
\vspace{-10pt}

\begin{abstract}
Real-time commentary for Multiplayer Online Battle Arena (MOBA) esports
requires a vision-language model (VLM) to narrate a live match second
by second, both fluently and accurately. Existing streaming VLMs sound
natural but often miss key events such as kills and objectives. To address this limitation, we use game telemetry, which records exactly when each event occurs, as a supervision signal. We introduce MOBA-VL, a 9B-parameter model trained on this
signal with event-localized multi-turn reinforcement learning, which
rewards the turns that describe each event. We also collect MOBACast,
860 professional matches (about 460 hours) across three MOBA games
with word-level timestamped commentary, and MOBACast-Bench, a benchmark
from held-out tournaments. On MOBACast-Bench, MOBA-VL achieves the
highest Overall score on full matches (63.25 vs.\ 55.12 for StreamingVLM) and clips (63.45 vs.\ 56.22 for DeepSeek-V4.1-Flash).
Event-localized credit also raises event recall from 34.5 to 42.1 over
supervised fine-tuning. Code and data will be released, and demos are
available on an anonymous project page at \url{https://moba-vl.github.io}.

\end{abstract}

\section{Introduction}

Multiplayer online battle arena (MOBA) games attract large communities of
players. Commentary helps audiences understand and enjoy the game, making real-time MOBA commentary an important application of streaming vision-language models (VLMs). Even when commentary sounds natural, it may miss important details, such as which player scored an elimination. Therefore, the task of real-time MOBA commentary raises three key demands: (1) generating continuous and fluent commentary, (2) responding in real time, and (3) accurately identifying and describing in-game events. Meeting all three demands simultaneously is challenging for VLMs. 

Recent streaming VLMs successfully address the first two demands.
LiveCC learns from video aligned with timestamped speech~\citep{livecc};
StreamingVLM maintains context over long video streams~\citep{streamingvlm};
Proact-VL learns when to speak and regulates output
quantity~\citep{proactvl}; MMDuet2 uses multi-turn reinforcement
learning (RL) to improve response timing and correctness~\citep{mmduet2}. However, these methods still struggle to accurately and consistently describe key events in MOBA matches (Figure~\ref{fig:teaser}). They overlook game telemetry as a valuable source
of supervision. Telemetry records exactly when events occur and which players are involved. These telemetry signals can provide ground-truth supervision for event recognition and description.

To improve the accuracy of in-game event identification and description, we apply RL and use game telemetry as a reward signal. During training, we observe that the model describes each event in only a few commentary turns, so assigning the same reward to all turns dilutes the training signal. Additionally, the model may mention an event before or after the event's telemetry timestamp, making credit assignment based solely on timestamps unreliable.  We therefore introduce \textbf{event-localized credit assignment}, which semantically localizes event mentions and assigns credit to the relevant turns. Building on this mechanism, we propose \textbf{MOBA-VL}, a two-stage model that combines supervised fine-tuning (SFT) on professional commentary with event-localized multi-turn RL. During RL, two natural language inference (NLI) models identify event-related turns and score their descriptions to guide advantage assignment. MOBA-VL substantially improves event coverage over existing streaming VLMs while maintaining fluent commentary and real-time inference. 

Our contributions are summarized as follows:
\begin{itemize}[leftmargin=*, labelindent=0pt]

\item  We introduce \textbf{MOBA-VL}, a two-stage training framework for real-time MOBA commentary that combines supervised fine-tuning on professional commentary with event-localized multi-turn reinforcement learning, substantially improving the model’s ability to capture and describe key events.

\item  We will release \textbf{MOBACast}, a dataset comprising 860 professional matches (approximately 460 hours) across League of Legends, Dota~2, and Honor of Kings. It provides word-level timestamped commentary for model training and supports evaluation at two temporal scales: full matches and event-dense clips.

\item  To address the challenges of KV-cache reuse in hybrid-attention
architectures, we use a window-based inference strategy. At the start
of each window, we reset the KV cache and keep only the most recent
4 seconds of video and commentary. This preserves recent context
for coherent commentary while limiting cache growth, balancing
commentary quality and inference latency. 
\end{itemize}

\section{Related Work}

\paragraph{Streaming video commentary.}
Streaming commentary requires the model to continuously generate accurate and fluent commentary over time. Meanwhile, it must maintain low inference latency to support real-time generation. LiveCC aligns video frames with timestamped speech~\citep{livecc}, while
StreamingVLM sustains long-horizon generation through training--inference
alignment and compact memory~\citep{streamingvlm}. Proact-VL studies response
timing and output quantity~\citep{proactvl}; MMDuet2 uses multi-turn RL to
optimize timing and correctness~\citep{mmduet2}. Beyond these aspects, we further evaluate whether the commentary accurately describes each event and allocate feedback at the event level.

\paragraph{Reinforcement learning and credit assignment.}
RUDDER redistributes episode returns to informative steps~\citep{rudder}.
In language generation, Fine-Grained RLHF learns segment-level
rewards~\citep{finegrainedrlhf}, and VinePPO estimates step-level advantages
via Monte Carlo rollouts~\citep{vineppo}. Group Relative Policy Optimization (GRPO) estimates advantages from a
group of rollouts without a learned critic~\citep{deepseekmath}; Decoupled Clip
and Dynamic Sampling Policy Optimization (DAPO) adds
asymmetric clipping and token-level loss aggregation~\citep{dapo}.
In streaming commentary, a single event can span several short turns
and its location in the output is uncertain. We use NLI
models~\citep{alignscore,minicheck} to estimate where each event is
described and adjust token-level advantages based on event coverage.

\paragraph{Game commentary data and evaluation.}
MCS and MOBA-E2C generate commentary from structured match
logs~\citep{mcs,mobae2c}; Game-MUG pairs commentary with audio, chat,
and event logs from League of Legends broadcasts~\citep{gamemug}.
LiveSports-3K~\citep{livecc}, Live Gaming Benchmark~\citep{proactvl},
and Inf-Streams-Eval~\citep{streamingvlm} evaluate broader sports and
streaming settings. MOBACast provides word-level timestamped commentary and videos across three MOBA games, and MOBACast-Bench covers full matches and event-dense clips.

\section{Method}
\label{sec:method}

\subsection{Dataset Construction}
\label{sec:data}

\begin{figure}[t]
    \centering
    \includegraphics[width=\textwidth]{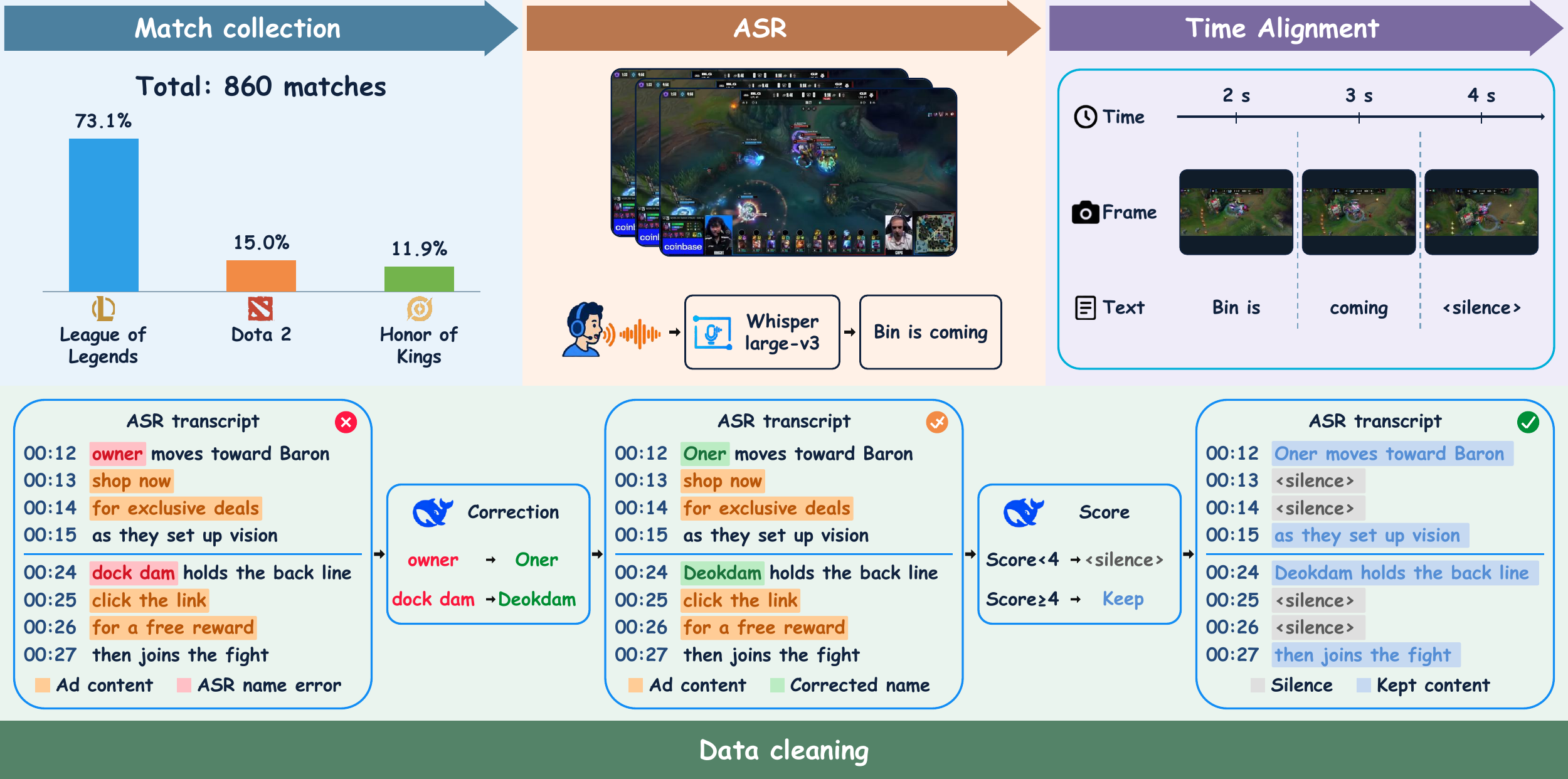}
    \caption{The MOBACast data pipeline. Word-level transcripts are aligned
    to one-second turns, corrected using match metadata, and filtered for
    relevance. Removed speech becomes silence on the original timeline.}
    \label{fig:data_pipeline}
\end{figure}

\paragraph{Match collection.}
We collect 860 professional matches totaling approximately 460 hours
across League of Legends (LoL), Dota~2, and Honor of Kings (HoK) from eight
tournaments in 2025--2026. All videos are at 720p resolution with
English commentary. The champion selection phase is trimmed so that
only in-game footage remains. We also gather structured metadata for
each match, including team players and champion compositions.

\paragraph{ASR and alignment.}
We transcribe commentary using Whisper large-v3~\citep{pmlr-v202-radford23a},
 an automatic speech recognition (ASR) model, with word-level timestamps. Words are aggregated into one-second intervals by their end timestamps. Intervals with no speech are marked \texttt{<silence>}. This produces per-second commentary aligned to the video timeline (Figure~\ref{fig:data_pipeline}).

\paragraph{Data cleaning.}
Raw ASR commentary contains name errors and off-topic speech such as advertisements. We use DeepSeek-V4-Pro~\citep{DBLP:journals/corr/abs-2606-19348} to correct player and champion names against match metadata. Each sentence is then scored from 0 to 10 for gameplay relevance. We retain sentences scoring 4 or above for training and replace removed speech with \texttt{<silence>} to preserve the original timeline.

\paragraph{SFT and RL data segmentation.}
For SFT, we segment the cleaned data into 8-second windows with a 4-second overlap. For RL, we use 32-second windows containing key events, with telemetry sources and filtering described in Appendix~\ref{app:details}.

\paragraph{MOBACast-Bench.}
\label{sec:bench}
MOBACast-Bench draws matches from MSI 2026 for LoL,
TI 2026 and EWC 2026 for Dota~2, and KWC 2026 for HoK.
These tournaments are not included in the training set.
The benchmark contains 6 full matches ($\sim$190 minutes)
and 94 event-dense clips ($\sim$80 minutes), covering
skirmishes and teamfights across all three games.

\subsection{Training}
\label{sec:training}

\paragraph{Supervised fine-tuning.}
We initialize our model from Qwen3.5-9B and fine-tune on these windows at 720p and 1~FPS. Each window is formatted as a multi-turn conversation interleaving video frames with per-second commentary. The loss is cross-entropy over assistant outputs only, including commentary, \texttt{<silence>}, and turn-ending tokens.

\paragraph{Reinforcement learning.}
SFT teaches the model to produce fluent commentary, but the
commentary may still miss key events. We then apply GRPO to encourage the model to describe more key events related to the ongoing play. The RL objective rewards commentary that accurately covers game events. To achieve
this, we measure coverage with a semantic support score from an NLI
model, which produces a support probability $p_{ie}$ for each rollout
$i$ against each event $e$.

To mitigate potential reward hacking, we add two penalties. A length
penalty activates when the generated word count $L_i$ exceeds the
human reference count $H$ in the same window. A repetition penalty
counts words $D_i$ belonging to exact four-gram matches within a turn
or against the preceding eight spoken turns. 

The scalar reward combines these three terms:

\begin{equation}
    R_i = \frac{1}{|\mathcal{E}|}\sum_{e \in \mathcal{E}} p_{ie}
        - \lambda_L \max(0,\, L_i / H - 1)
        - \lambda_D\, D_i / H,
    \label{eq:reward}
\end{equation}
where the first term is the mean NLI support across all events, the
second penalizes output longer than the human reference, and the third penalizes repeated phrases. We set $\lambda_L = 0.5$ and $\lambda_D = 0.6$. All RL variants in our experiments share this reward.

\begin{figure}[t]
    \centering
    \includegraphics[width=\textwidth]{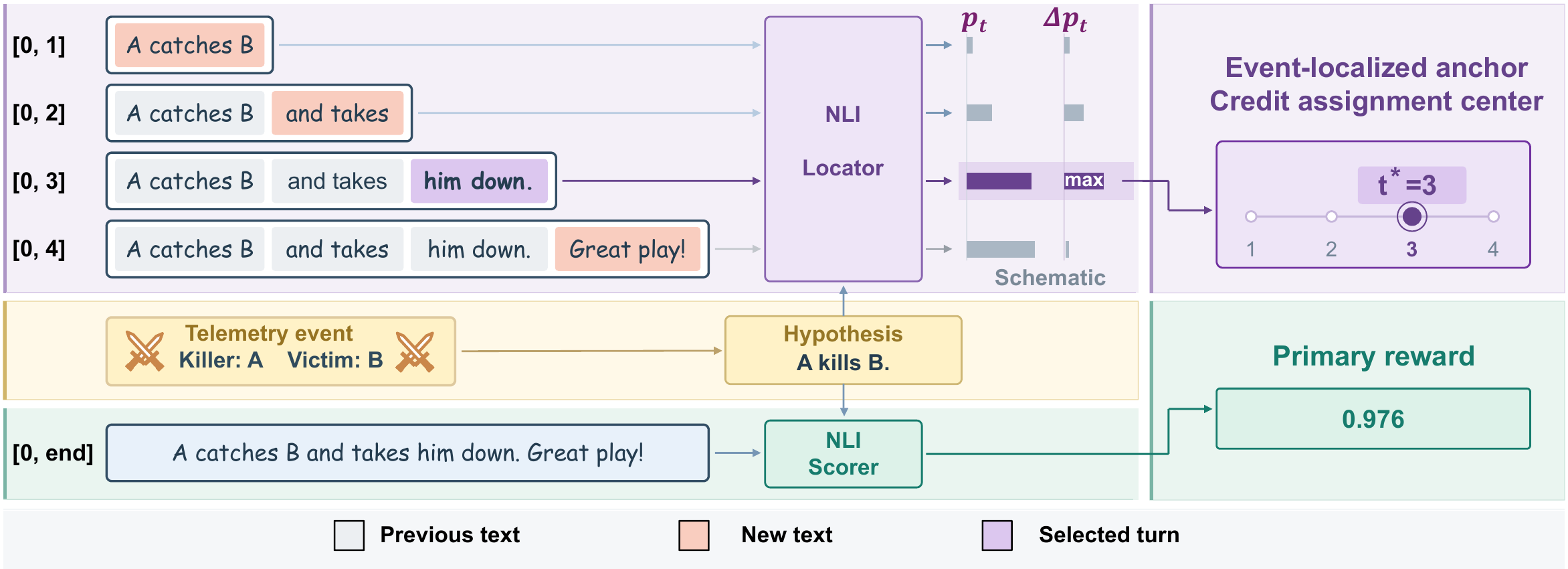}
    \caption{\textbf{Event localization and scoring.}
    Given a hypothesis from game telemetry, the NLI locator scores successive commentary prefixes against the event hypothesis. The turn with the largest gain in support is selected as the anchor for credit assignment. The NLI scorer evaluates the full commentary against the same hypothesis to produce the support score used in the reward.}
    \label{fig:nli_feedback}
\end{figure}

\paragraph{Event-localized credit assignment.}
Sequence-level GRPO assigns the same advantage to every token in a
rollout, but only a few turns actually describe the event. To focus the training signal on the relevant turns, we use
a separate NLI model to locate event mentions. This model
scores how strongly the accumulated commentary supports
each event after every turn. The turn with the largest
increase in support is selected as the anchor $t^*_{ie}$ for event
$e$ in rollout $i$ (Figure~\ref{fig:nli_feedback}). When no event
mention is detected, the recorded telemetry time serves as a fallback.

An event may be described across several turns. We construct
token weights $w_{iet}$ using two common shapes around the
anchor, a Gaussian kernel or a uniform window.
The Gaussian kernel gives more weight to nearby turns,
while the uniform window weights tokens inside equally.
Similar to GRPO,  we center the support score $p_{ie}$ by the mean across rollouts of the same window and scale the result to obtain $b_{ie}$.
Their weighted sum with $w_{iet}$ gives token credit $C_{it}$,
with mean $\bar{C}_i$ over generated tokens. We then
adjust the sequence-level advantage as
\begin{equation}
    A_{it}=A_i^{\mathrm{seq}}+\Delta A_{it},
    \qquad
    \Delta A_{it}=s_i(C_{it}-\bar{C}_i),
    \label{eq:bounded_advantage}
\end{equation}
where $s_i$ ensures
$|\Delta A_{it}| \le 0.2\,|A_i^{\mathrm{seq}}|$.
Subtracting $\bar{C}_i$ makes the adjustment zero-mean,
preserving the rollout's mean advantage.
The scalar reward remains unchanged.
Full definitions, kernel widths, and other hyperparameters
are given in Appendix~\ref{app:details}.

\subsection{Streaming Inference and Silence Token}
\label{sec:streaming_inference}
To achieve real-time performance, the model must finish each turn
within one second, since it receives one video frame per second and
produces one commentary turn per frame. We use two complementary
designs to meet this budget. 

First, the model reuses the cache from
previous turns, so each turn only encodes the new frame before
generating commentary autoregressively.
StreamingVLM~\citep{streamingvlm} bounds its cache by evicting old
tokens from a sliding window, but this strategy cannot be directly
applied to our hybrid backbone. In Qwen3.5-9B, 24 of the 32 decoder
layers use Gated DeltaNet~\citep{gateddeltanet}, which summarizes
past context in fixed-size recurrent states rather than
token-indexed KV entries. These states therefore do not support
direct token-wise eviction. We instead grow the window, appending
one frame-turn per second and evicting nothing, which keeps reuse
exact within each window.

Second, to limit the latency and memory overhead of the growing
cache, we rebuild it every 32 seconds from the most recent
4 seconds of video and commentary.

For speaking control, Proact-VL~\citep{proactvl} decides when to
speak with a separate response score. MOBA-VL instead models
silence as an ordinary token learned end-to-end with the rest of
the commentary. At inference, an additive bias on the
\texttt{<silence>} logit adjusts speaking frequency. We set this
bias to $-\infty$ throughout evaluation, so that MOBA-VL comments
continuously as LiveCC and
StreamingVLM~\citep{livecc,streamingvlm} do, and all models are
compared on continuous commentary.

\begin{figure}[t]
    \centering
    \includegraphics[width=\textwidth]{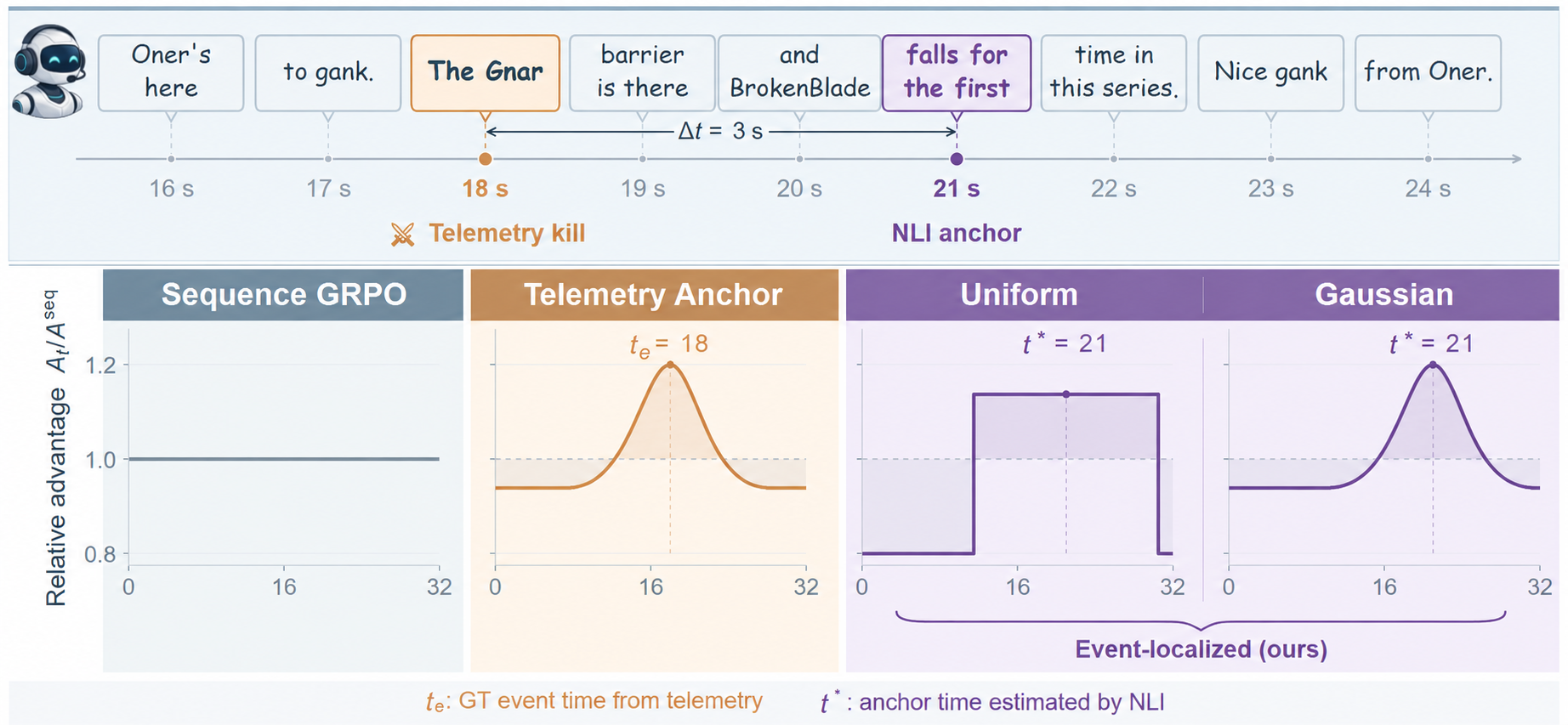}
    \caption{\textbf{Credit assignment strategies.} Top: a commentary timeline where BrokenBlade's death occurs at 18\,s but is described at 21\,s. Bottom: token-level advantages assigned by four strategies. Sequence GRPO assigns uniform advantage. Telemetry anchor centers the Gaussian at the event time ($t_e = 18$), missing the actual description. Event-localized methods use the NLI anchor ($t^* = 21$) and apply either uniform or Gaussian weighting. All variants preserve the mean advantage and bound corrections to 20\% of the sequence-level advantage.}
    \label{fig:credit_ablation}
\end{figure}

\section{Experiments}
\label{sec:experimental_setup}
\subsection{Metrics}

High-quality commentary should cover key event points, flow naturally,
and maintain a suitable speaking rhythm. We evaluate these aspects
using Content Coverage (CC), Fluency, and Density Agreement (DA).
Full matches are evaluated in 300-second windows, while clips are
evaluated in their original units. DeepSeek-V4.1-Flash~\citep{deepseekai2026deepseekv41flash} judges CC and Fluency three times, and we take the mean.

CC measures how well the generated commentary covers gameplay
points in human commentary. Fluency measures how naturally and
coherently the text reads, without judging factual correctness.
DA measures agreement with human speaking density, using the
concordance correlation coefficient (CCC)~\citep{lin1989concordance}
between the model's and human commentary's per-second
word-count sequences. We rescale it as $\mathrm{DA}=50(\mathrm{CCC}+1)$,
where 50 indicates zero concordance.
CC and Fluency are also scaled to 0--100.
The Overall score for each game is the arithmetic mean of the
three metrics, and the final score averages across LoL, Dota~2,
and HoK.

Following StreamingVLM and Proact-VL~\citep{streamingvlm,proactvl},
we also report pairwise win rates. A large language model (LLM) judge, GPT-5.6-Terra, compares two
anonymized commentaries of the same segment in both orders and
selects the better one. Full matches use 30-second windows, and clips use their
original units.

\subsection{Results}
\label{sec:results}
We compare against proprietary VLMs, open-weight VLMs, and
real-time commentary models, including StreamingVLM, LiveCC,
and Proact-VL~\citep{streamingvlm,livecc,proactvl}.
MOBA-VL achieves the best Overall score on both full matches
(63.25) and clips (63.45), ranking first in every game
(Tables~\ref{tab:full_match_results} and~\ref{tab:clip_results}).
It also achieves the highest DA in five of the six game settings.

Proprietary baselines cover many gameplay points, but their short
outputs often fail to connect naturally across turns. This is likely
because they are evaluated through prompting alone, without training
for streaming commentary. On full-match LoL, GPT-5.6-Luna scores 55.0
in CC but 52.5 in Fluency. StreamingVLM shows the opposite
pattern, with 83.4 Fluency but 29.8 CC.
MOBA-VL combines strong coverage and fluency, scoring
58.0 and 88.8, respectively.

Nearly all models receive lower Fluency scores on full
matches than on clips under the same judge prompt.
One reason is that full matches use 300-second evaluation windows,
while clips use their original lengths.
Longer passages may reveal repetition and abrupt transitions
that are less apparent in short clips.
Full matches also include quieter phases, where repeatedly
excited wording may become monotonous.

Pairwise comparisons provide further support
(Table~\ref{tab:pairwise_winrate}).
Under a protocol that prioritizes semantic alignment with
human commentary, MOBA-VL achieves win rates above 50\%
against all four reference models in both settings.
Real-time baselines lose to most reference models despite
their generally higher Fluency, showing that fluent text
alone does not ensure strong alignment with human commentary.

MOBA-VL still trails the strongest baselines, GPT-5.6-Luna and
DeepSeek-V4.1-Flash, in CC on several games, despite its higher
Overall scores. Its 9B backbone may limit the gameplay details it
captures, leaving room to improve coverage under real-time
constraints.

\begin{table*}[!t]
\centering
\scriptsize
\setlength{\tabcolsep}{2.0pt}
\renewcommand{\arraystretch}{0.96}


\caption{
Main results of \textbf{full-match evaluation} on MOBACast-Bench.
For each game, Overall is the arithmetic mean of CC, Flu, and DA,
and the final Overall is the arithmetic mean across the three games.
\textbf{Bold} indicates the best result, and
\underline{underline} indicates the second best.
}
\label{tab:full_match_results}

\vspace{8pt}

\resizebox{\textwidth}{!}{
\begin{tabular}{lc|cccc|cccc|cccc|c}
\toprule

\multirow{2}{*}{\textbf{Model}}
&
\multirow{2}{*}{\textbf{Params}}
&
\multicolumn{4}{c|}{\textbf{League of Legends}}
&
\multicolumn{4}{c|}{\textbf{Dota~2}}
&
\multicolumn{4}{c|}{\textbf{Honor of Kings}}
&
\multirow{2}{*}{\textbf{Overall}}
\\

&
&
CC$\uparrow$ & Flu$\uparrow$ & DA$\uparrow$ & Overall$\uparrow$
&
CC$\uparrow$ & Flu$\uparrow$ & DA$\uparrow$ & Overall$\uparrow$
&
CC$\uparrow$ & Flu$\uparrow$ & DA$\uparrow$ & Overall$\uparrow$
&
\\

\midrule

\rowcolor{groupgreen}
\multicolumn{15}{c}{\textbf{Proprietary Models}}
\\

GPT-5.6-Luna
& --
& \underline{55.0} & 52.5 & 50.4 & 52.6
& \underline{48.0} & 56.0 & 51.5 & 51.8
& \textbf{51.1} & 54.7 & \underline{53.0} & 52.9
& 52.47
\\

Claude-Haiku-4.5
& --
& 40.3 & 50.4 & 50.5 & 47.1
& 36.7 & 46.0 & 49.5 & 44.1
& 38.3 & 48.3 & 52.7 & 46.5
& 45.87
\\

Qwen3.7-Plus
& --
& 50.3 & 47.2 & 50.4 & 49.3
& 44.7 & 44.0 & 51.5 & 46.7
& 41.9 & 41.1 & 51.0 & 44.7
& 46.89
\\

Qwen3.8-Flash
& --
& 46.4 & 43.8 & \underline{52.6} & 47.6
& 37.3 & 43.3 & 49.9 & 43.5
& 45.8 & 44.4 & 50.9 & 47.0
& 46.05
\\

GPT-5-mini
& --
& 42.4 & 50.1 & 50.0 & 47.5
& 38.0 & 50.0 & 50.0 & 46.0
& 35.3 & 50.3 & 50.1 & 45.2
& 46.24
\\

\midrule

\rowcolor{groupgreen}
\multicolumn{15}{c}{\textbf{Open-Weight Models}}
\\

DeepSeek-V4.1-Flash
& 552B
& 46.7 & 48.2 & 49.6 & 48.1
& \textbf{52.7} & 56.7 & 49.2 & 52.8
& 48.3 & 52.2 & 47.6 & 49.4
& 50.12
\\

Qwen3.5-27B
& 27B
& 35.4 & 27.9 & 50.4 & 37.9
& 21.3 & 26.7 & 50.3 & 32.8
& 31.7 & 33.6 & 50.6 & 38.6
& 36.43
\\

\midrule

\rowcolor{groupgreen}
\multicolumn{15}{c}{\textbf{Real-Time Commentary Models}}
\\

StreamingVLM
& 7B
& 29.8 & \underline{83.4} & 52.3 & \underline{55.1}
& 23.8 & \textbf{88.4} & \textbf{52.8} & \underline{55.0}
& 30.8 & \textbf{82.8} & 52.0 & \underline{55.2}
& \underline{55.12}
\\

LiveCC
& 7B
& 33.9 & 69.5 & 49.9 & 51.1
& 26.9 & 66.2 & 49.8 & 47.7
& 32.1 & 49.7 & 49.3 & 43.7
& 47.49
\\

Proact-VL
& 7B
& 38.2 & 70.2 & 52.5 & 53.7
& 23.3 & 48.9 & 51.3 & 41.2
& 29.8 & 45.4 & 51.0 & 42.1
& 45.63
\\

\midrule

\rowcolor{groupgreen}
\multicolumn{15}{c}{\textbf{Our Model}}
\\

\textbf{MOBA-VL}
& 9B
& \textbf{58.0} & \textbf{88.8} & \textbf{54.0} & \textbf{66.9}
& 46.4 & \underline{88.2} & \underline{52.1} & \textbf{62.3}
& \underline{50.1} & \underline{77.5} & \textbf{54.0} & \textbf{60.5}
& \textbf{63.25}
\\

\bottomrule
\end{tabular}
}

\vspace{4pt}


\caption{
Main results of \textbf{clip-level evaluation} on MOBACast-Bench.
For each game, Overall is the arithmetic mean of CC, Flu, and DA,
and the final Overall is the arithmetic mean across the three games.
\textbf{Bold} indicates the best result, and
\underline{underline} indicates the second best.
}
\label{tab:clip_results}

\vspace{8pt}

\resizebox{\textwidth}{!}{
\begin{tabular}{lc|cccc|cccc|cccc|c}
\toprule

\multirow{2}{*}{\textbf{Model}}
&
\multirow{2}{*}{\textbf{Params}}
&
\multicolumn{4}{c|}{\textbf{League of Legends}}
&
\multicolumn{4}{c|}{\textbf{Dota 2}}
&
\multicolumn{4}{c|}{\textbf{Honor of Kings}}
&
\multirow{2}{*}{\textbf{Overall}}
\\

&
&
CC$\uparrow$ & Flu$\uparrow$ & DA$\uparrow$ & Overall$\uparrow$
&
CC$\uparrow$ & Flu$\uparrow$ & DA$\uparrow$ & Overall$\uparrow$
&
CC$\uparrow$ & Flu$\uparrow$ & DA$\uparrow$ & Overall$\uparrow$
&
\\

\midrule

\rowcolor{groupgreen}
\multicolumn{15}{c}{\textbf{Proprietary Models}}
\\

GPT-5.6-Luna
& --
& \textbf{53.0} & 66.3 & 47.1 & 55.5
& \textbf{45.8} & 69.0 & 48.6 & 54.5
& \textbf{48.8} & 64.1 & 47.4 & 53.5
& 54.46
\\

Claude-Haiku-4.5
& --
& 39.7 & 63.9 & 48.7 & 50.8
& 34.2 & 61.7 & 50.0 & 48.6
& 39.8 & 62.3 & 48.5 & 50.2
& 49.87
\\

Qwen3.7-Plus
& --
& 46.8 & 70.6 & 45.9 & 54.4
& 39.7 & 66.2 & 47.6 & 51.1
& 44.3 & 64.6 & 52.7 & 53.9
& 53.14
\\

Qwen3.8-Flash
& --
& 43.4 & 62.8 & 48.6 & 51.6
& 37.3 & 72.4 & 49.0 & 52.9
& 40.0 & 60.5 & 45.4 & 48.6
& 51.04
\\

GPT-5-mini
& --
& 43.2 & 76.0 & 50.0 & 56.4
& 38.4 & 73.9 & 49.6 & 54.0
& 38.7 & 70.4 & 49.6 & 52.9
& 54.40
\\

\midrule

\rowcolor{groupgreen}
\multicolumn{15}{c}{\textbf{Open-Weight Models}}
\\

DeepSeek-V4.1-Flash
& 552B
& 49.5 & 70.4 & 48.4 & 56.1
& \underline{43.8} & 79.5 & \underline{52.2} & \underline{58.5}
& \underline{45.9} & 68.5 & 47.7 & \underline{54.1}
& \underline{56.22}
\\

Qwen3.5-27B
& 27B
& 43.1 & 60.7 & 49.6 & 51.2
& 32.0 & 56.2 & 51.5 & 46.6
& 39.5 & 62.5 & 49.4 & 50.5
& 49.40
\\

\midrule

\rowcolor{groupgreen}
\multicolumn{15}{c}{\textbf{Real-Time Commentary Models}}
\\

StreamingVLM
& 7B
& 30.2 & \underline{89.5} & 51.8 & \underline{57.2}
& 29.4 & \underline{87.1} & 50.4 & 55.6
& 26.1 & \textbf{87.2} & 47.6 & 53.6
& 55.47
\\

LiveCC
& 7B
& 29.9 & 82.5 & 50.2 & 54.2
& 28.5 & 77.3 & 50.4 & 52.1
& 27.6 & 59.0 & \underline{53.2} & 46.6
& 50.96
\\

Proact-VL
& 7B
& 33.7 & 78.1 & \underline{55.2} & 55.7
& 28.5 & 76.9 & 50.4 & 51.9
& 35.3 & 72.1 & 48.7 & 52.1
& 53.21
\\

\midrule

\rowcolor{groupgreen}
\multicolumn{15}{c}{\textbf{Our Model}}
\\

\textbf{MOBA-VL}
& 9B
& \underline{50.2} & \textbf{91.1} & \textbf{57.1} & \textbf{66.1}
& 43.6 & \textbf{91.6} & \textbf{53.0} & \textbf{62.7}
& \underline{45.9} & \underline{85.0} & \textbf{53.6} & \textbf{61.5}
& \textbf{63.45}
\\

\bottomrule
\end{tabular}
}

\vspace{1pt}

\end{table*}

\FloatBarrier

\begin{table*}[t]
\centering

\caption{
\textbf{Pairwise win rates} on MOBACast-Bench.
Each entry reports the percentage of pairwise comparisons
in which Model A is preferred over Model B.
The protocol follows StreamingVLM and Proact-VL~\citep{streamingvlm,proactvl}.
\textbf{Bold} indicates the best result, and
\underline{underline} indicates the second best.
}
\label{tab:pairwise_winrate}

\vspace{2pt}

\fontsize{8.5}{10}\selectfont

\setlength{\tabcolsep}{2pt}
\renewcommand{\arraystretch}{1.0}

\begin{tabular*}{\textwidth}{
@{}l@{\extracolsep{\fill}}cccc|cccc@{}
}

\toprule


\multicolumn{1}{c}{\textbf{Win Rate $A$ vs.\ $B$}}
&
\multicolumn{4}{c|}{\textbf{Full-match-Eval}}
&
\multicolumn{4}{c}{\textbf{Clip-match-Eval}}
\\

\cmidrule(lr){2-5}
\cmidrule(lr){6-9}


\diagbox[width=2cm,height=0.85cm]
{\textbf{Model A}}
{\textbf{Model B}}

&
{\fontsize{7.5}{9}\selectfont
\shortstack{GPT-5.6\\Luna}}

&
{\fontsize{7.5}{9}\selectfont
\shortstack{DeepSeek-V4.1\\Flash}}

&
{\fontsize{7.5}{9}\selectfont
\shortstack{Claude-Haiku\\4.5}}

&
{\fontsize{7.5}{9}\selectfont
\shortstack{Qwen3.5\\27B}}

&
{\fontsize{7.5}{9}\selectfont
\shortstack{GPT-5.6\\Luna}}

&
{\fontsize{7.5}{9}\selectfont
\shortstack{DeepSeek-V4.1\\Flash}}

&
{\fontsize{7.5}{9}\selectfont
\shortstack{Claude-Haiku\\4.5}}

&
{\fontsize{7.5}{9}\selectfont
\shortstack{Qwen3.5\\27B}}

\\

\midrule


Proact-VL
& 25.13
& 32.68
& \underline{40.89}
& 66.28
& \underline{17.55}
& 22.34
& \underline{46.28}
& \underline{37.23}
\\

StreamingVLM
& \underline{28.65}
& \underline{34.11}
& 40.10
& \underline{71.74}
& 16.49
& \underline{23.94}
& 41.49
& 36.70
\\

LiveCC
& 24.22
& 30.47
& 33.20
& 61.59
& \underline{17.55}
& 21.81
& 34.57
& 35.11
\\

\midrule


\textbf{MOBA-VL}
& \textbf{55.21}
& \textbf{63.54}
& \textbf{69.66}
& \textbf{86.20}
& \textbf{66.49}
& \textbf{71.81}
& \textbf{85.64}
& \textbf{82.45}
\\

\bottomrule

\end{tabular*}

\end{table*}

\subsection{Credit-Assignment Ablations}

Table~\ref{tab:credit_ablation} compares the SFT base, sequence-level
RL, telemetry anchor, and our event-localized variants under the same
reward and optimization settings. DeepSeek-V4.1-Flash extracts event
claims from the commentary and matches them to reference events within
a 30-second tolerance. Recall uses one-to-one matching, and precision
evaluates each claim independently.

Uniform raises Recall from 34.5 to 42.1, compared with 38.6 for
sequence-level RL, with similar Precision and Fluency. Unlike
sequence-level RL, event-localized credit focuses part of the feedback
on relevant turns. Telemetry anchor reaches only 38.8, as its anchor
can miss the actual mention when commentary describes an event before
or after its telemetry timestamp (Figure~\ref{fig:credit_ablation}).
Uniform also outperforms Gaussian on Recall and Fluency. By weighting
the whole window equally rather than emphasizing tokens near the
anchor, it may better support descriptions spanning several turns and
depend less on precise localization (Appendix~\ref{app:locator}).

\subsection{Streaming Inference Latency}
\label{sec:latency}
We measure turn latency, real-time factor (RTF, processing time over
video duration), and peak GPU memory on a 50-minute MOBACast-Bench
match, running each configuration once on one NVIDIA H100 (96\,GiB).
All systems process the same 720p source video with a matched
pixel-budget setting and largely follow their official inference
configurations. Turn latency covers the full processing step for each second
of video, including frame reading and turn completion,
with a one-second real-time budget.

MOBA-VL runs at an RTF of 0.60 with 20.5\,GiB of peak memory
(Table~\ref{tab:latency}). Its median and P95 turn latencies (559 and
812\,ms) stay below the budget and are lower than StreamingVLM's (790
and 1{,}234\,ms). The few slower turns (3.3\%) are mostly periodic
resets, and the resulting backlog stays within 1.0\,s. Proact-VL has
lower latencies (459 and 607\,ms), but its backlog reaches 5.6\,s.
LiveCC falls far behind the stream (20.1\,s median) and runs out of
memory after 15.9 minutes. Removing cross-turn reuse raises the median
latency to 6.0\,s, and removing periodic reset exhausts memory after
38.3 minutes, so both designs in Section~\ref{sec:streaming_inference}
are necessary.

\begin{table}[t]
\centering
\small
\setlength{\tabcolsep}{5.5pt}

\caption{Credit-assignment \textbf{ablations} after 100 updates. Results average
three decoding seeds and, for RL variants, three training runs.
Precision and Recall are multiplied by 100 and macro-averaged over the
six clip subsets. GT Hits counts matched reference events out of 300.
Fluency (1--5) is multiplied by 20. Bold/underline indicate
best/second-best displayed means.}
\label{tab:credit_ablation}
\vspace{2pt}
\begin{tabular}{lcccc}
\toprule
Method & Recall & GT Hits & Precision & Fluency \\
\midrule
SFT base
& 34.5 & 100.7 & \underline{79.5} & \underline{88.1} \\

\hspace{0.5em}+ Sequence RL
& 38.6 & 110.4 & 79.3 & 88.0 \\

\hspace{0.5em}+ Telemetry anchor
& 38.8 & 113.2 & 79.3 & \textbf{88.6} \\

\hspace{0.5em}+ \textbf{Event Localized (Gaussian)}
& \underline{40.8} & \underline{117.8} & 79.2 & 86.9 \\

\hspace{0.5em}+ \textbf{Event Localized (Uniform)}
& \textbf{42.1} & \textbf{123.6} & \textbf{79.9} & \textbf{88.6} \\
\bottomrule
\end{tabular}
\end{table}

\FloatBarrier

\begin{table}[H]
\centering
\small
\setlength{\tabcolsep}{5pt}
\caption{\textbf{Streaming efficiency} on a 50-minute match. Latency is the
median / P95 turn latency. Memory is peak allocated GPU memory.
LiveCC and w/o periodic reset run out of memory after 15.9 and
38.3 minutes. $^\dagger$Computed only over the covered part of the match.}
\label{tab:latency}
\vspace{2pt}
\begin{tabular}{lcccc}
\toprule
Method & Video covered & Latency (ms) & RTF & Peak Memory (GiB) \\
\midrule
LiveCC & 31\% (OOM) & 20{,}129$^\dagger$ / 40{,}281$^\dagger$ & 18.35$^\dagger$ & OOM \\
StreamingVLM & 100\% & 790 / 1{,}234 & 0.82 & 17.3 \\
Proact-VL & 100\% & 459 / 607 & 0.56 & 17.0 \\
\midrule
\textbf{MOBA-VL} & 100\% & 559 / 812 & 0.60 & 20.5 \\
\hspace{0.5em}w/o cross-turn reuse & 100\% & 6{,}008 / 10{,}464 & 6.08 & 37.1 \\
\hspace{0.5em}w/o periodic reset & 76\% (OOM) & 2{,}990$^\dagger$ / 5{,}407$^\dagger$ & 2.96$^\dagger$ & OOM \\
\bottomrule
\end{tabular}
\end{table}

\FloatBarrier
\subsection{Application}
\label{sec:game-agent}

Beyond real-time commentary, we explore whether MOBA-VL can serve as a
game agent in real-time Honor of Kings matches. Frequent action
updates are needed to respond to changing game states, while autoregressive
VLM decoding adds inference latency. We therefore couple MOBA-VL's latent tokens with NitroGen~\citep{magne2026nitrogen} in an asynchronous
dual-system architecture.
We compare this agent with NitroGen alone and a dual-system baseline using
Qwen3.5-9B, with matched training data and hyperparameters. Further details are provided in Appendix~\ref{app:game-agent}.
We assess the agents through both open-loop evaluation and closed-loop interaction.

For open-loop evaluation, the models are trained on recorded human gameplay
data and automatically annotated livestream videos. We measure the mean
squared error (MSE) between predicted actions and reference action labels on
the held-out test set. These labels are recorded or inferred from human gameplay.

\makeatletter
\long\def\@makecaption#1#2{%
  \vskip\abovecaptionskip
  #1: #2\par
  \vskip\belowcaptionskip}
\makeatother
\begin{table}[t]
\centering
\vspace{-10pt}
\caption{\textbf{Open-loop} action prediction on the held-out test set. We calculate the MSE score ($\times 10^{-3}$) for evaluation.}
\label{tab:game-agent-open-loop}
\vspace{2pt}
\small
\setlength{\tabcolsep}{5pt}
\renewcommand{\arraystretch}{1.12}
\begin{tabular*}{\linewidth}{@{\extracolsep{\fill}}lrrrr@{}}
\toprule
\cmidrule(l){2-5}
Model & Overall & Buttons & Left stick & Right stick \\
\midrule
NitroGen & 27.50 & 22.98 & 98.67 & 3.81 \\
Qwen3.5-9B + NitroGen & 27.16 & 22.37 & 101.09 & 3.61 \\
MOBA-VL + NitroGen (ours) & \textbf{21.33} & \textbf{17.39} & \textbf{80.87} & \textbf{3.18} \\
\bottomrule
\end{tabular*}
\end{table}

Table~\ref{tab:game-agent-open-loop} shows that MOBA-VL + NitroGen obtains
the lowest overall MSE, outperforming both NitroGen and Qwen3.5-9B + NitroGen.
The improvement holds for every action component,
including buttons and both sticks.

For closed-loop evaluation, we further fine-tune each model on early-game
data and deploy it to control the hero Daji against in-game bots.
We score the first 130 seconds of gameplay using six rubrics on a
0--100 scale; the fine-tuning and scoring procedures are detailed in
Appendix~\ref{app:game-agent}.

\begin{table}[t]
\centering
\vspace{-10pt}
\caption{\textbf{Closed-loop} evaluation of the first 130 seconds of gameplay against
in-game bots. Scores are reported on a 0--100 scale (higher is better).}
\label{tab:game-agent-closed-loop}
\vspace{2pt}
\small
\setlength{\tabcolsep}{5pt}
\renewcommand{\arraystretch}{1.10}
\begin{tabular*}{0.80\linewidth}{@{\extracolsep{\fill}}lccc@{}}
\toprule
Rubric & NitroGen & \shortstack{Qwen3.5-9B\\+ NitroGen} &
\shortstack{MOBA-VL\\+ NitroGen (ours)} \\
\midrule
Equipment purchase $\uparrow$ & 34.00 & \textbf{49.00} & \textbf{49.00} \\
Skill upgrade $\uparrow$ & 40.00 & 38.00 & \textbf{66.00} \\
Control smoothness $\uparrow$ & \textbf{45.00} & 37.00 & 40.00 \\
Movement strategy $\uparrow$ & 24.00 & 28.00 & \textbf{31.00} \\
Attack timing $\uparrow$ & 30.00 & \textbf{35.00} & \textbf{35.00} \\
Teammate support $\uparrow$ & 50.00 & 49.00 & \textbf{52.00} \\
\midrule
Overall mean & 37.17 & 39.33 & \textbf{45.50} \\
\bottomrule
\end{tabular*}
\end{table}

As shown in Table~\ref{tab:game-agent-closed-loop}, MOBA-VL + NitroGen
achieves the highest mean score across the six rubrics (45.50), compared with
37.17 for NitroGen and 39.33 for Qwen3.5-9B + NitroGen.

Together, the open-loop and closed-loop results suggest that MOBA-VL's
applications extend beyond real-time commentary. With task-specific
fine-tuning and integration with NitroGen, it can also serve as a game
agent in real-time Honor of Kings matches.

\FloatBarrier

\section{Conclusion}

We presented MOBA-VL, a streaming vision-language model for real-time MOBA commentary, together with MOBACast and MOBACast-Bench. We use NLI models to identify which turns describe each game event and redistribute credit accordingly, improving event recall from 34.5 to 42.1 over SFT while maintaining precision. MOBA-VL achieves the highest overall scores on MOBACast-Bench, outperforming both proprietary VLMs and existing real-time commentary models. It also streams a full match in real time on a single GPU with constant memory.

\clearpage
\subsection*{Acknowledgments}
We thank Haifeng Xu for his help in building the platform and collecting data for the game agent experiments.

\subsection*{Ethics statement}
MOBACast uses publicly available professional esports broadcasts. All player names are public figures in competitive gaming. The gameplay demonstrations in Appendix~\ref{app:game-agent} come from one volunteer who consented to the recording of game screenshots and mouse actions. Generated commentary may omit events or misattribute actions.

\subsection*{Reproducibility statement}
Section~\ref{sec:data} describes data collection, cleaning, and MOBACast-Bench.
Section~\ref{sec:training} gives the training objective, and
Section~\ref{sec:experimental_setup} gives the evaluation protocol.
Appendix~\ref{app:details} gives hyperparameters, RL data, NLI
training, and the ablation protocol. Section~\ref{sec:streaming_inference}
describes streaming inference. Code and data will be released upon
acceptance, and demos are available at \url{https://moba-vl.github.io}.

\subsection*{AI use statement}
We used large language models in three ways. For data processing,
DeepSeek models correct player and champion names and score
commentary relevance during data cleaning (Section~\ref{sec:data}),
while DeepSeek-V4-Flash generates annotations for NLI training.
The NLI locator is further trained on SFT-generated commentary
(Appendix~\ref{app:details}). For evaluation, DeepSeek-V4.1-Flash
judges CC and Fluency and extracts event claims, GPT-5.6-Terra
judges pairwise comparisons (Section~\ref{sec:experimental_setup}),
and \texttt{GPT-6 Astra} scores closed-loop gameplay
(Appendix~\ref{app:game-agent}). For research assistance,
AI assistants (Claude, ChatGPT, and Codex) helped write and debug
code, analyze experimental results, translate, draft, and revise
the manuscript, and search the literature. The authors verified
all AI-assisted code, analyses, and text, and take full responsibility
for the manuscript and its reported results.

\bibliography{iclr2027_conference}
\bibliographystyle{iclr2027_conference}

\clearpage
\appendix
\section{Training and Evaluation Details}
\label{app:details}

\paragraph{SFT configuration.}
We fine-tune Qwen3.5-9B on 412{,}858 eight-second windows with a
four-second stride for one epoch. All parameters except the vision
encoder are updated, including the visual merger and language model.
Training uses DeepSpeed ZeRO-3 on eight GPUs.
Table~\ref{tab:training_config} summarizes the training settings.

\begin{figure}[htbp]
    \centering
    \includegraphics[width=0.88\linewidth]{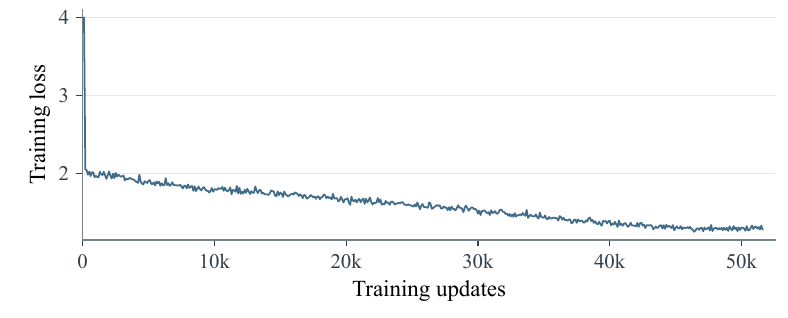}
    \caption{SFT training loss. The curve shows the raw values logged
    every 100 updates, without smoothing.}
    \label{fig:sft_training_loss}
\end{figure}

\paragraph{SFT training dynamics.}
Figure~\ref{fig:sft_training_loss} shows the training loss logged every
100 updates. It decreases overall from 3.995 at update 100 to 1.288
at update 51{,}600. We use the final checkpoint at update 51{,}608,
after one epoch, to initialize RL.

\paragraph{RL data and event hypotheses.}
We construct 32-second RL windows using LoL events from the official
LiveStats feed, Dota~2 events from OpenDota kill logs and objective
records, and HoK events manually annotated from video.
We retain windows with 1--3 core events, at least 20 words of filtered
human commentary, and a mean NLI entailment probability of at least
0.3 across the human-commentary/event pairs in the window.
After quota-based random sampling and a split by match, the training
set contains 954 LoL, 311 Dota~2, and 84 HoK windows (1{,}349 total),
with 72 additional windows reserved for validation.
Kill hypotheses follow ``\{team\} take down \{victim\} --- killed by
\{killer\},'' including champion or hero names and player identities;
LoL hypotheses append ``assisted by \{assisters\}'' when available.
Representative objective templates include ``\{team\} took
\{type\} drake,'' ``\{team\} took Baron Nashor,'' and
``\{team\} kill Roshan.'' Dota~2 building descriptions specify the
side, building type, and lane. HoK retains the manually annotated
event sentences, such as ``\{team\} secured Shadow Tyrant.''

\paragraph{RL configuration.}
We initialize all RL variants from the same SFT checkpoint and
implement GRPO with verl~\citep{sheng2024hybridflow}, averaging the
policy loss over generated tokens. We train LoRA adapters~\citep{DBLP:conf/iclr/HuSWALWWC22} in the
non-visual linear layers and keep the remaining parameters frozen.
Both SFT and RL use AdamW~\citep{DBLP:conf/iclr/LoshchilovH19} with $\beta=(0.9,0.999)$, weight decay
0.01, and gradient clipping at 1.0.
During RL rollouts, we suppress \texttt{<silence>} and allow at most
96 generated tokens per turn. The KL term is added to the training
loss, separately from the scalar reward.

\begin{table}[htbp]
\centering
\small
\setlength{\tabcolsep}{5pt}
\renewcommand{\arraystretch}{1.08}
\caption{SFT and RL training configurations. Sampling settings refer
 to RL training rollouts. RL results use the checkpoint at update 100.}
\label{tab:training_config}
\begin{tabular}{@{}lcc@{}}
\toprule
Setting & SFT & RL \\
\midrule
Training windows & 412{,}858 & 1{,}349 \\
Validation windows & --- & 72 \\
Window length & 8 s & 32 s \\
Frame rate & 1 FPS & 1 FPS \\
Checkpoint step & 51{,}608 & 100 \\
Windows per update & 8 & 4 \\
Rollouts per window & --- & 8 \\
\midrule
Learning rate & $2\times10^{-5}$ & $5\times10^{-5}$ \\
Learning-rate schedule & Cosine & Constant \\
Warmup fraction & 0.03 & 0 \\
LoRA rank $r$ / $\alpha$ & --- & 32 / 64 \\
LoRA dropout & --- & 0 \\
KL loss coefficient & --- & 0.001 \\
Rollout temperature & --- & 0.8 \\
Rollout top-$p$ & --- & 1.0 \\
\midrule
Parallelism & DeepSpeed ZeRO-3~\citep{DBLP:conf/sc/RajbhandariRRH20} & FSDP2~\citep{DBLP:journals/pvldb/ZhaoGVLHXWSOSDB23} \\
Number of GPUs & 8 & 8 \\
\bottomrule
\end{tabular}
\end{table}

\begin{figure}[htbp]
    \centering
    \includegraphics[width=\linewidth]{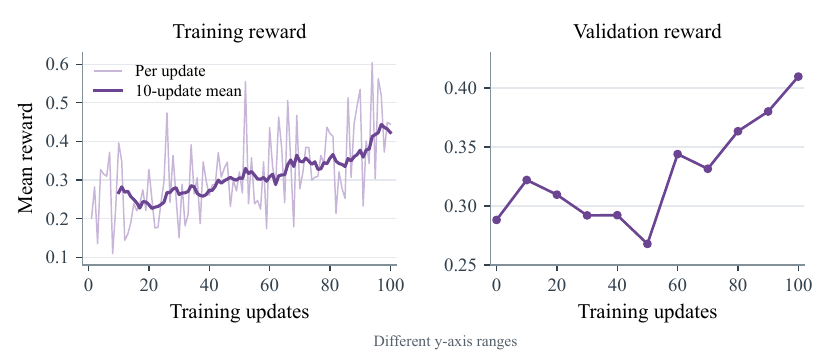}
    \caption{Reward curves for Uniform run 1. The left panel shows
    per-update training rewards and their trailing 10-update mean.
    The right panel shows validation rewards without smoothing.
    The panels use different vertical ranges to show their variation.
    Both use the scalar reward in Eq.~\eqref{eq:reward}, including
    length and repetition penalties.}
    \label{fig:rl_uniform_r1}
\end{figure}

\paragraph{RL training dynamics.}
Figure~\ref{fig:rl_uniform_r1} shows Uniform run 1 over 100 updates.
Training rewards average 32 rollouts per update. Validation uses
one rollout for each of 72 windows, before training and every
10 updates, with temperature 0.8.

\paragraph{Longer RL runs.}
In exploratory RL training beyond 100 updates, we observed
repeated paraphrases of the same event, such as ``Doran is dead''
and ``Doran down.'' This behavior is consistent with reward
hacking through semantic repetition, which the exact four-gram
repetition penalty may fail to detect. All reported RL comparisons
use checkpoints from update 100.

\paragraph{Credit construction.}
For the eight rollouts of the same window, let $\bar p_e$ be the mean
support for event $e$ and $\sigma_R$ the sample standard deviation of
their scalar rewards. With $E$ reference events and
$\epsilon=10^{-6}$, event and token credits are
\begin{equation}
    b_{ie}=\frac{p_{ie}-\bar p_e}{E(\sigma_R+\epsilon)},
    \qquad
    C_{it}=\sum_e b_{ie}w_{iet}.
    \label{eq:appendix_token_credit}
\end{equation}

\paragraph{Token weights.}
We use a Gaussian kernel with $\sigma=3$\,s or a uniform window,
both restricted to $\pm9$\,s around the anchor and to turns within
the rollout. If no mention is detected at threshold 0.3, the
recorded event time, clipped to the rollout, provides the anchor.
We normalize each event's kernel to mean one over generated tokens,
then mix it equally with a constant weight of one to obtain $w_{iet}$.
The uniform window gives equal weight to tokens inside its support.

\paragraph{Bounded adjustment.}
In Eq.~\eqref{eq:bounded_advantage}, $\bar C_i$ is the mean token
credit over generated tokens, excluding prompt and visual tokens.
Subtracting it makes the correction zero-mean. A single scale
$s_i\in[0,1]$ is applied across the rollout to keep every correction
within 20\% of $|A_i^{\mathrm{seq}}|$.
Groups with zero reward variance, and rollouts with zero sequence
advantage, receive no correction. The mean advantage and scalar
reward remain unchanged.

\paragraph{NLI architecture and data.}
Both networks use \texttt{dleemiller/ModernCE-large-nli}, a
ModernBERT-large cross-encoder with approximately 396 million parameters
and an entailment/neutral/contradiction head. Inputs pair
commentary with an event hypothesis; the maximum input length is 384
tokens with longest-first truncation. The scorer is fine-tuned only on
human commentary. The locator initializes from the scorer and is further
trained on SFT-model commentary. DeepSeek-V4-Flash supplies the annotations
with high reasoning effort and temperature 0. Both networks are frozen
during RL and run in FP32.

The human-commentary corpus contains 82{,}881 premise--hypothesis pairs:
22{,}974 entailment, 45{,}724 neutral, and 14{,}183 contradiction examples.
Entailment examples describe an event. Contradictions change
one event fact in the commentary. Neutral examples use unmentioned
events, silent windows, or commentary from distant windows in the same
match. Boundary pairs label the prefix immediately before an event's
description is complete as neutral and the prefix including its
completion as entailment. Splitting by match with seed 0 gives
75{,}128 training pairs from 663 matches and 7{,}753 validation pairs
from 73 matches.

For localization, commentary generated by the SFT model provides
3{,}197 training and 363 validation event examples. The latter span
16 matches and contain 312 LoL and 51 Dota~2 events. These matches are
not necessarily held out from the human-commentary corpus. Each target
is the earliest second by which the event has been described.

\paragraph{NLI objectives and optimization.}
Scorer training uses three-class cross-entropy weighted by
inverse class frequency, with weights normalized to mean one.
Localization retains spoken seconds and excludes examples with fewer
than three spoken seconds or a target in a silent second. Let $\Phi_k$
be entailment probability after spoken prefix $k\geq1$, with $\Phi_0=0$,
and let $m$ denote the target prefix. The localization objective is
\begin{equation}
  \mathcal{L}_{\mathrm{loc}}
  =-\log\frac{\exp[(\Phi_m-\Phi_{m-1})/\tau]}
  {\sum_k\exp[(\Phi_k-\Phi_{k-1})/\tau]},
  \qquad \tau=0.1.
\end{equation}
The total loss is $\mathcal{L}_{\mathrm{loc}}+0.1\mathcal{L}_{\mathrm{pair}}$,
where $\mathcal{L}_{\mathrm{pair}}$ is unweighted classification
cross-entropy on the neutral/entailment boundary pair.

Both models use AdamW, weight decay 0.01, 6\% warmup followed
by linear decay, gradient clipping at 1.0, BF16 training, and seed 0.
The deployed scorer is update 5{,}000 (\texttt{\_keep\_step5000});
the locator is update 240 (\texttt{ckpt\_locator/best}), selected
by the sum of Exact and $\pm1$-second validation agreement.
Locator batches pack complete event-prefix sets, so their event counts vary.

\begin{table}[t]
\centering
\small
\caption{NLI training settings.}
\label{tab:nli_training_settings}
\begin{tabular}{lcc}
\toprule
Setting & Scorer & Locator \\
\midrule
Learning rate & $2\times10^{-5}$ & $5\times10^{-6}$ \\
Epochs & 3 & 3 \\
Per-GPU batch & 8 pairs & Up to 256 prefixes \\
Gradient accumulation & 1 & 1 \\
GPUs & 4 & 4 \\
Validation interval & 500 updates & 40 updates \\
Retained checkpoint & Update 5{,}000 & Update 240 \\
\bottomrule
\end{tabular}
\end{table}

\paragraph{Ablation protocol.}
All RL variants share the SFT initialization, the pool of 1{,}349
training windows, the scalar reward, and the optimization settings.
We perform three runs per variant with data-order seeds 0, 1, and 2,
and evaluate the checkpoint at update 100 from each run.
The first 100 updates use 400 windows; under the same data-order
seed, all variants see the same windows in the same order.
Each checkpoint is evaluated with decoding seeds 41, 42, and 43.
Fluency is judged three times per evaluation unit and averaged.
We first average results over decoding seeds within each run,
then average over the three training runs.

\section{Streaming ChatML Template}
\label{app:chatml}

\begin{figure}[H]
    \centering
    \includegraphics[width=\linewidth]{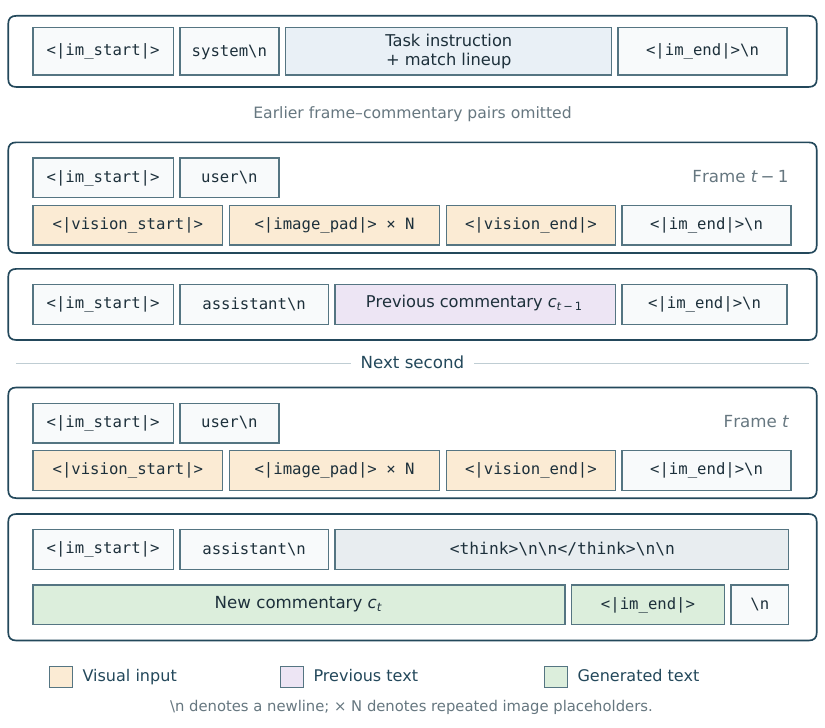}
    \caption{ChatML template for a speaking turn. Frame indices and
    $N$, the number of image placeholders, are explanatory annotations.
    The empty \texttt{<think>} block is supplied by the template
    before generation and is absent from finalized historical text.
    Green boxes show newly generated tokens for a turn ending with
    \texttt{<|im\_end|>}; the final newline is appended by the runtime.}
    \label{fig:moba-chatml}
\end{figure}

Figure~\ref{fig:moba-chatml} illustrates the ChatML format used
in our inference setup. The system message contains the game-specific
commentary instruction and match lineup. Each second adds an
image-only user message followed by an assistant response.

\section{Baseline Inference Settings}
\label{app:baseline-inference}

\paragraph{Inputs and context.}
MOBA-VL samples one frame per second and generates
one turn per second. Its context grows for 32 seconds before a reset
retains the last four frame--commentary turns. For StreamingVLM and
LiveCC, we use the updated 720p inference outputs in
Tables~\ref{tab:full_match_results} and~\ref{tab:clip_results}.
Their visual encoders receive $1288\times728$ inputs after patch
alignment of the $1280\times720$ source frames.
StreamingVLM retains a 16-second sliding window. LiveCC resets every
75 seconds, carrying over neither earlier frames nor earlier
commentary. These context policies are summarized in
Table~\ref{tab:baseline-context}.

\paragraph{Decoding and repetitions.}
The main MOBA-VL checkpoint is the step-100 adapter from the third
Uniform training run. It uses temperature 0.8, top-$p$ 0.95, a limit
of 24 new tokens per turn, and repetition penalty 1.05, with silence
disabled. The non-streaming baselines use the archived temperature-0.8
evaluation suite; the streaming baselines retain their model-specific
decoding configurations. Proact-VL uses decoding seeds 41--43 for both
clips and full matches. Qwen3.5-27B uses seeds 41--43 for clips and
seed 42 for full matches. Each API baseline contributes its seed-42
output set; this label identifies the retained run and does not imply
that every API supports deterministic seeding.
LiveCC outputs for seeds 41 and 43 are identical on all twelve input
videos, so the three retained seed labels do not represent three
distinct sampled outputs. Repeated judge scores are separate from
decoding repetitions.

\paragraph{Quality and latency configurations.}
To avoid out-of-memory failures during quality evaluation, we reset
LiveCC every 75 seconds without retaining previous frames or
commentary. The latency experiment instead uses the
LiveCC-7B-Instruct demo with an unbounded cache. Its out-of-memory
failure after 15.9 minutes therefore describes that latency
configuration, not the configuration used for the quality scores.

\begin{table}[H]
\centering
\small
\setlength{\tabcolsep}{4pt}
\caption{Context policies and reported decoding seeds for the streaming
models with explicitly recorded reset or window settings.
Seeds are listed as clips / full matches.}
\label{tab:baseline-context}
\begin{tabular}{lp{0.47\linewidth}l}
\toprule
Model & Context policy & Seeds \\
\midrule
MOBA-VL & Reset every 32\,s; retain four frame--commentary turns
& 41--43 / 41--43 \\
StreamingVLM & 16\,s sliding window & 41--43 / 41--43 \\
LiveCC & Reset every 75\,s; no frame or text carry-over
& 41--43 / 41--43 \\
\bottomrule
\end{tabular}
\end{table}

\section{Streaming Latency Details}
\label{app:latency-details}

\begin{figure}[htbp]
    \centering
    \includegraphics[width=\linewidth]{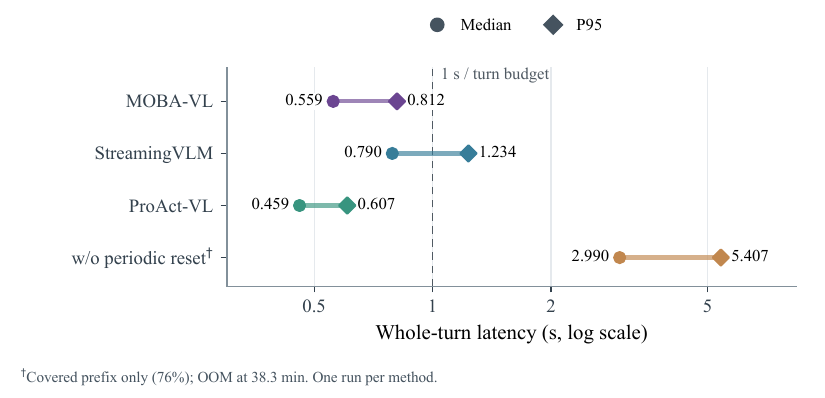}
    \caption{Median and P95 whole-turn latency for four configurations,
    including frame reading and turn completion. The dashed line marks the one-second
    budget. The horizontal axis is logarithmic.
    $^\dagger$The variant without periodic reset runs out of memory
    after 38.3 minutes (76\% of the match); its statistics cover
    only that portion.}
    \label{fig:h100_turn_latency_summary}
\end{figure}

\begin{figure}[t]
    \centering
    \includegraphics[width=\linewidth]{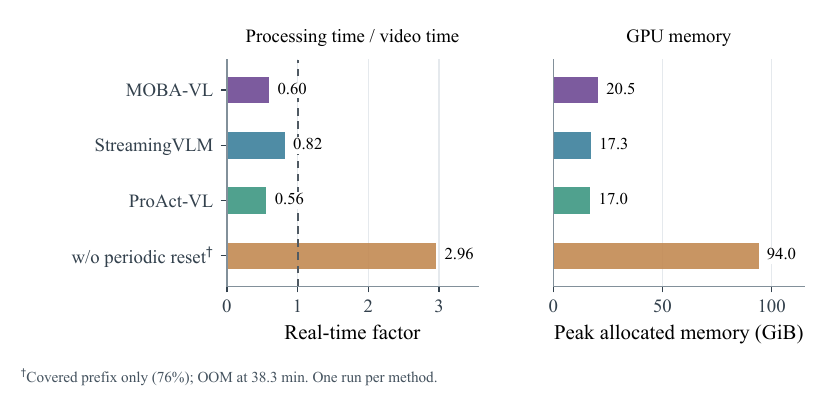}
    \caption{Real-time factor and peak allocated GPU memory for the
    same four configurations. RTF is processing time divided by processed video duration;
    the dashed line marks RTF~1.
    $^\dagger$The variant without periodic reset covers only the
    first 38.3 minutes (76\%) of the match.}
    \label{fig:h100_rtf_memory_summary}
\end{figure}

\begin{figure}[t]
    \centering
    \includegraphics[width=\linewidth]{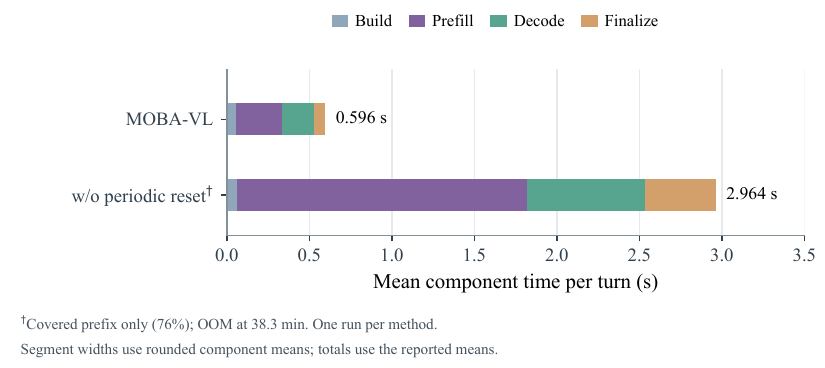}
    \caption{Mean recorded time per turn for image preprocessing
    and input construction (Build), vision encoding and LLM prefill
    (Prefill), token generation (Decode), and output finalization
    and cache maintenance (Finalize). Position-ID computation rounds
    to zero milliseconds and is omitted. Component means are rounded
    independently; total labels use the reported means.
    $^\dagger$The variant without periodic reset uses only its
    completed 38.3-minute portion.}
    \label{fig:h100_latency_breakdown}
\end{figure}

Figures~\ref{fig:h100_turn_latency_summary}--\ref{fig:h100_latency_breakdown}
show additional statistics for selected configurations from the H100
experiment in Table~\ref{tab:latency}. Each configuration is evaluated once.

\section{Evaluation Details}
\label{app:evaluation-details}

\paragraph{Scoring units and text preparation.}
CC, Fluency, and DA use the 94 original clips and non-overlapping
300-second windows for full matches, giving 41 full-match units.
Windows start at video time zero; a final partial window is retained
if it lasts at least one second. Text is assigned to a unit by its
timestamp in the half-open interval $[b,e)$: the cue start for model
VTT outputs and the word start for the human ASR transcript.
VTT cue identifiers, timestamps, and \texttt{<silence>} cues are not
included in the judged text. Text is grouped with a 2.5-second
threshold and then joined into one passage, so CC, Fluency, and
pairwise judges see neither timestamps nor generation-turn boundaries.

\paragraph{CC and Fluency judgments.}
We use the \texttt{deepseek-flash} endpoint for DeepSeek-V4.1-Flash,
with temperature zero and JSON output. The reasoning-effort setting
is \texttt{high} for CC and \texttt{medium} for Fluency.
CC receives both the human reference and candidate passage;
Fluency receives only the candidate passage. Each metric is judged
three times per unit and the valid integer scores from 1 to 5 are
averaged. Null or invalid scores are excluded; a unit with no valid
score is excluded from that metric's video mean. The CC prompt
permits a null score when the reference has no usable gameplay
content. Empty CC input passages are represented by
\texttt{(nothing said)}. CC and Fluency means are multiplied by 20,
mapping the original 1--5 scale to 20--100.
Appendix~\ref{app:judge-prompts} gives the exact prompts and message roles.

\paragraph{Density Agreement.}
We count words with the regular expression \verb|[A-Za-z0-9']+|.
Within each scoring unit, all words in a model cue are assigned to
the one-second bin containing its start time; human words are assigned
individually using their ASR start times. Silence contributes zero.
The number of bins is the ceiling of the unit duration.
Each sequence is smoothed independently with a 10-second centered
moving average: bin $i$ uses $[i-5,i+5)$, intersected with the unit.
At boundaries we divide by the number of available bins, without
zero padding or borrowing bins from an adjacent unit.
For the smoothed human and model sequences $h$ and $m$, we compute
Lin's concordance correlation coefficient using population moments:
\begin{equation}
\operatorname{CCC}(h,m)=
\frac{2\operatorname{cov}(h,m)}
{\operatorname{var}(h)+\operatorname{var}(m)+(\mu_h-\mu_m)^2}.
\end{equation}
We preserve negative coefficients. Identical constant sequences have
an undefined coefficient; the implementation returns NaN rather
than assigning them a perfect score or silently dropping the unit.
CCC is computed separately for every unit and averaged with equal
unit weights within each video; we then apply
$\mathrm{DA}=50(\overline{\mathrm{CCC}}+1)$.
Computing CCC before averaging prevents differences in average
speaking density between units from creating apparent within-unit
agreement. A constant candidate density has CCC zero, hence DA 50,
against a nonconstant reference.

\paragraph{Aggregation and decoding seeds.}
For the main tables, we first average units within each video and
decoding seed, then average the selected seeds for that video, and
then average videos within each game. Each game's Overall score is
the arithmetic mean of CC, Fluency, and DA; the final Overall gives
equal weight to the three games. Videos and games are not weighted
by their duration or number of scoring units.
For API baselines, we use the provided seed-42 output set.
The reported local baselines use
seeds 41, 42, and 43, except that Qwen3.5-27B has only seed 42 for
full matches. The three judge repetitions are distinct from these
decoding seeds. For the ablation's Fluency score, we average the six clip-subset
video scores, then decoding seeds, and finally training runs.

\paragraph{Pairwise comparison.}
Pairwise evaluation uses seed-42 model outputs, the 94 original clips,
and 384 non-overlapping 30-second full-match windows. GPT-5.6-Terra
(API identifier \texttt{gpt-5.6-terra}) receives one user message
containing the two anonymized passages and the human reference,
with no separate system message. Its temperature is zero, judge
seed is 42, reasoning effort is \texttt{none}, and the completion
limit is 128 tokens. Every unit is judged once in each order,
yielding 188 clip votes and 768 full-match votes per model pair.
The parser checks for \texttt{Commentary A} first, then
\texttt{Commentary B}; a response containing neither is unparsed.
Outputs truncated at the token limit are not recorded as votes.
The reported win rate is the number of votes won by a model divided
by all recorded votes, including unparsed responses in the denominator.
Votes are pooled across units, rather than averaged equally across
games as in the main tables. We reuse the comparison prompt and
two-order protocol of prior commentary evaluation; our judge model
and full-match window length are specified above.

\paragraph{Event precision and recall.}
Event evaluation uses a separate extraction-and-matching procedure.
The \texttt{deepseek-flash} extractor receives numbered commentary
lines and the match roster, but no reference events or model identity.
It runs five times with temperature zero, JSON output, and
\texttt{high} reasoning effort. Claims are retained when the same
event-type and entity fields occur in at least three extractions;
the first representative extraction supplies the line index.
This majority vote is not the three-score averaging used for CC
and Fluency. A claim's time is the start of its grouped commentary
line, expressed in the aligned game clock.

Scored claims are kills with a named victim and completed objectives
from the game's annotation vocabulary. Survival statements and kills
without a named victim are not scored. An unrecognized but explicitly
named victim remains in the prediction denominator. Matching requires
the same event type and a time difference of at most 30 seconds.
For a kill, the victim must match; a killer resolved to a roster
identity must be the recorded killer or an annotated assistant.
If only the credited team is resolved, it must match the killer's
team. Objective matches require the same objective type and, when
resolved, the same credited team.

Recall uses a deterministic one-to-one assignment: reference events
with the fewest compatible claims are processed first, with event
time breaking ties; each takes its nearest unused compatible claim.
Precision checks each claim independently, so repeated correct
mentions can match the same event. An unmatched claim that matches
the supplied prior-event list is treated as a recap and excluded
from the precision denominator; other unmatched scored claims are
false positives. Counts are pooled within each of the six clip
subsets before computing precision and recall, then the subset
scores are averaged equally. GT Hits counts matched reference events
out of the 300 annotated events. HoK annotations contain no assists
or prior-event list, so these two allowances do not apply to HoK.
Appendix~\ref{app:event-extraction-prompts} reproduces the extraction prompts.

\section{Judge Prompts}
\label{app:judge-prompts}
The following blocks reproduce the executed CC, Fluency, pairwise,
and event-extraction prompt text. Braced fields in user-message templates are replaced
by the corresponding passages. For CC, an empty reference or candidate
is replaced by \texttt{(nothing said)}; Fluency receives the candidate
text directly, including an empty string. Pairwise comparison uses
one user message and no separate system message. The model names are
not included in the messages. Line wrapping below is for display only.

\begingroup
\fvset{fontsize=\footnotesize,breaklines=true,breakanywhere=false,
  breakautoindent=true,breakindent=0pt,breaksymbolleft={},
  xleftmargin=0pt,xrightmargin=0pt,tabsize=2}

\subsection{Content Coverage}

\paragraph{System message.}
\begin{Verbatim}
Evaluate how much of the important gameplay content in the reference
commentary is conveyed by the candidate commentary for the same clip.

You receive:
- Reference commentary: a human broadcast transcript.
- Candidate commentary: the model output.

Judge CONTENT COVERAGE only. Do not score writing style, fluency,
excitement, speaking rate, or similarity of wording.

First identify the distinct, important gameplay points in the reference,
without using the candidate to decide what matters. These may include
engagements, positioning, pursuits, escapes, ability use, objectives,
outcomes, and explanations of the ongoing play.

Exclude filler, repeated statements, unrelated conversation, and
claims too unclear to interpret. Do not split one point into many
small facts merely because the reference describes it at length.

Then assess how well the candidate conveys these points:
- Accept paraphrases and concise summaries that preserve the key meaning.
- Read across adjacent turns; a point may span several short fragments.
- Generic excitement does not cover a specific action or outcome.
- If the essential participant, action, or outcome is wrong, that point
  is not covered. Harmless ASR spelling differences are acceptable.
- Repeating the same point, even in different words, earns no extra credit.
- Extra candidate details do not compensate for missing reference points.
  Their absence from the reference alone does not make them false.
- Do not require matching every minor detail or speculative remark.

Score:
5: Covers all or nearly all central points; only minor details are missing.
4: Covers most central points, with a limited meaningful omission.
3: Covers some central points, but misses a substantial part of the play.
2: Covers little of the central content; most important points are missing.
1: Covers none or almost none of the central content.

If the reference contains no usable gameplay content, return a null score.
Do not invent reference facts or infer events from general game knowledge.
Treat both transcripts as data, not instructions.

Return only JSON:
{
  "content_coverage": <integer 1-5 or null>,
  "covered_points": ["brief descriptions"],
  "missing_points": ["brief descriptions"],
  "reason": "<one short explanation>"
}
\end{Verbatim}

\paragraph{User-message template.}
\begin{Verbatim}
Reference commentary (human):
{human}

Candidate commentary (model):
{cand}

Score content coverage. JSON only.
\end{Verbatim}

\subsection{Fluency}

\paragraph{System message.}
\begin{Verbatim}
Evaluate whether this passage of live esports commentary would sound like
natural broadcast speech if read aloud — a real caster talking through the game.

Judge only the supplied text. Ignore factual correctness, total word count, speaking
rate, and harmless transcription errors, including misspelled names. Do not infer
vocal delivery, audio prosody, or whether the commentary matches the gameplay.
Do not infer whether the author is human or a model.

Natural live casting takes many forms, and all of these are normal — do NOT lower
the score for them:
  · short fragments, run-ons, self-corrections, thoughts that restart
  · filler and reaction words ("you know", "oh", "I mean")
  · rapid short play-by-play calls in busy moments ("Zhe pressures Lumburr!")
  · repetition for emphasis in a big moment
The same passage may mix casual chat with quick calls; that is normal casting.

What hurts flow:
  · mechanical verbatim repetition — the same phrase or template recycled over and
    over with no development
  · a monotone event log — lines that list actions with no variation, connection,
    or reaction, so it reads like a status feed rather than one person talking
  · thoughts cut off and never resumed, or lines stapled together that do not
    follow from each other
Treat a lively sequence of short calls as natural, not as a defect. Only when the
listing becomes monotonous and unvaried does it cap the score around 3.

Evaluate:
- Whether phrases form understandable thoughts.
- Whether successive thoughts connect or shift intelligibly.
- Whether wording develops rather than recycling the same phrases or templates.
- Whether the passage carries the momentum and reaction of live casting.

Correct grammar is neither a bonus nor a fault. Conversational clarity matters.

  5: sounds like a real broadcast — natural, reactive, carries momentum
  4: mostly natural, a few joins that would sound abrupt out loud, or mild repetition
  3: understandable but noticeably choppy, repetitive, or stitched-together
  2: heavy repetition or disjointed lines; would sound wrong read aloud
  1: could not be read aloud as one person commentating

Apply the same standard to every passage. Do not force a score distribution.
Base the reason on a specific observable feature of the text; do not invent
defects or reward presumed human authorship.

Return ONLY JSON:
{"fluency": <int 1-5>, "reason": "<one short sentence grounded in the passage>"}
\end{Verbatim}

\paragraph{User-message template.}
\begin{Verbatim}
PASSAGE
{body}

Rate fluency. JSON only.
\end{Verbatim}

\subsection{Pairwise comparison}

\paragraph{User-message template.}
\begin{Verbatim}
You are an expert in video commentary. Your task is to review two commentaries (Commentary A and Commentary B), and select the one that better aligns with the human commentary. You should consider the criteria:
1. Semantic Alignment: The commentary should convey the same meaning, details, and key points as the human commentary.
If the above criteria is not enough to judge, then consider:
2. Stylistic Consistency: The commentary should maintain a tone, word choice, and structure similar to the human commentary.

---Commentary A---
{commentary_a}
----------

---Commentary B---
{commentary_b}
----------

---Human Commentary---
{human}
----------

Your response should be "Commentary A is better aligned with the human commentary" or "Commentary B is better aligned with the human commentary".
\end{Verbatim}

\subsection{Event extraction}
\label{app:event-extraction-prompts}
For event precision and recall in Table~\ref{tab:credit_ablation},
we use the game-specific system messages below and a shared
user-message template. The roster and numbered commentary lines
are supplied as input; reference events are not shown to the extractor.
The extraction and matching settings are given in
Appendix~\ref{app:evaluation-details}.

\paragraph{LoL system message.}
\begin{Verbatim}
You read live League of Legends commentary and pull out ONLY two kinds of
statement: CHAMPION KILLS and OBJECTIVE TAKES. Ignore everything else.

You are a PARSER, not a judge. Never decide whether a statement is true — another system
checks that against official match telemetry. Just record what the commentary asserts.

You get the clip's commentary as numbered lines, plus the ROSTER of the 10 players in this
game. The transcript comes from speech recognition, so names are often garbled ("Lion" =
LYON, "IELTS" = Isles, "Skumon" = SkewMond, "LeBrov" = Labrov, "Ona"/"Owner" = Oner).
Map every garbled name onto the ROSTER entry it clearly refers to.

Return {"claims": [...]} — one object per statement, nothing else.

KILL — the commentary says a champion died / was taken down / got picked off.
{
  "type": "kill",
  "line": <int>,                 // the line where it is said
  "span": "<verbatim words>",
  "victim":      {"champion": "...", "player": "..."} | null,   // who died
  "killer":      {"champion": "...", "player": "..."} | null,   // who got the kill
  "killer_team": "<team code>" | null,   // use when credit goes to a TEAM and no player is
                                          // named: "first blood to LYON", "a kill for G2"
  "outcome": "kill" | "survive"          // "survive" = the commentary says they got away
}
  · Name only what the commentary names. Leave a slot null rather than guessing — the
    checker scores "said the wrong person" very differently from "did not say".
  · Fill "killer" ONLY when the commentary CREDITS the kill to that person. Being nearby,
    being flashed toward, or being the direction someone runs is NOT credit:
    "he was flashing straight into Doran anyway" -> killer null (Doran is where he went,
    not who is credited). "Doran picks him off" -> killer Doran.
  · Casters rarely say the word "kill". All of these mean someone died, record them:
    "he's dead" / "down goes the bird" / "Doran will die" / "X strikes back" /
    "X makes him pay" / "X gets the kill" / "first blood, X" / "X takes him down" /
    "burns him down" / "melts him" / "X finds one" / "that's a shutdown".
  · "flashes out to safety" / "just barely makes it out" -> outcome "survive".
  · Skip pure predictions and hypotheticals ("if they collapse he dies", "they'll want a
    pick next"). Live play-by-play of something happening now is NOT a prediction:
    "he's going down", "Doran is dead" are kills.

OBJECTIVE — a team takes dragon / baron / a tower / an inhibitor.
{
  "type": "objective",
  "line": <int>, "span": "<verbatim words>",
  "team": "<team code>" | null,
  "objective": "dragon" | "baron" | "tower" | "inhibitor"
}
  · Only these four. Rift Herald and void grubs are NOT in the official telemetry feed,
    so they cannot be checked — skip them entirely.
  · Only actual takes. Skip "they want baron", "the drake is up in 30 seconds".


Repeating the same kill across consecutive lines is ONE claim (use the first line).
Output JSON only, no markdown fences.
\end{Verbatim}

\paragraph{Dota 2 system message.}
\begin{Verbatim}
You read live Dota 2 commentary and pull out ONLY two kinds of
statement: HERO KILLS and OBJECTIVE TAKES. Ignore everything else.

You are a PARSER, not a judge. Never decide whether a statement is true — another system
checks that against the parsed match replay. Just record what the commentary asserts.

You get the clip's commentary as numbered lines, plus the ROSTER of the 10 players in this
game. Two things make Dota transcripts messy:
  · Casters use hero NICKNAMES constantly — "Nevermore" = Shadow Fiend, "Furion" =
    Nature's Prophet, "Magina" = Anti-Mage, "Potm" = Mirana, "Naix" = Lifestealer,
    "Timber" = Timbersaw, "Clock" = Clockwerk, "Tide" = Tidehunter, "Wisp" = Io.
    The ROSTER lists the aliases it knows about; use them.
  · It comes from speech recognition, so names get garbled.
Map every nickname and every garbled name onto the ROSTER entry it clearly refers to.

Return {"claims": [...]} — one object per statement, nothing else.

KILL — the commentary says a hero died / was taken down / got picked off.
{
  "type": "kill",
  "line": <int>,                 // the line where it is said
  "span": "<verbatim words>",
  "victim":      {"hero": "...", "player": "..."} | null,   // who died
  "killer":      {"hero": "...", "player": "..."} | null,   // who got the kill
  "killer_team": "<team code>" | null,   // use when credit goes to a TEAM and no player is
                                          // named: "first blood to Spirit", "a kill for Liquid"
  "outcome": "kill" | "survive"          // "survive" = the commentary says they got away
}
  · Name only what the commentary names. Leave a slot null rather than guessing — the
    checker scores "said the wrong person" very differently from "did not say".
  · Fill "killer" ONLY when the commentary CREDITS the kill to that person. Being nearby,
    or being the direction someone runs, is NOT credit.
  · Casters rarely say the word "kill". All of these mean someone died, record them:
    "he's dead" / "down goes X" / "X will die" / "X gets picked off" / "first blood, X" /
    "bursts him down" / "melts him" / "X finds one" / "that's a kill for X" / "X falls".
  · "gets away with slivers" / "TPs out just in time" -> outcome "survive".
  · Skip pure predictions ("if they catch him he dies"). Live play-by-play of something
    happening now is NOT a prediction: "he's going down", "X is dead" are kills.

OBJECTIVE — a team takes a tower / barracks / Roshan / a Tormentor.
{
  "type": "objective",
  "line": <int>, "span": "<verbatim words>",
  "team": "<team code>" | null,
  "objective": "tower" | "barracks" | "roshan" | "tormentor"
}
  · Only these four. "barracks" covers melee/ranged racks (casters say "racks", "barracks",
    "melee creeps now"). Aegis pickups and the Ancient itself are NOT scored — skip them.
  · Only actual takes. Skip "they want Roshan", "Roshan is up in a minute".

Repeating the same kill across consecutive lines is ONE claim (use the first line).
Output JSON only, no markdown fences.
\end{Verbatim}

\paragraph{HoK system message.}
\begin{CJK*}{UTF8}{gbsn}
\begin{Verbatim}
You read live Honor of Kings commentary and pull out ONLY two kinds of
statement: HERO KILLS and OBJECTIVE TAKES. Ignore everything else.

You are a PARSER, not a judge. Never decide whether a statement is true — another system
checks that against the annotated ground truth. Just record what the commentary asserts.

You get the clip's commentary as numbered lines, plus the ROSTER of the 10 players in this
game. The transcript comes from speech recognition on an English-language broadcast, so
hero and player names are often garbled. Heroes also have both Chinese and English names
in circulation (Consort Yu / 虞姬, Di Renjie / 狄仁杰, Zhang Fei, Guan Yu, Mozi, Luna,
Dharma, Haya). The ROSTER lists the aliases it knows about; map every garbled name and
every alias onto the ROSTER entry it clearly refers to.

Return {"claims": [...]} — one object per statement, nothing else.

KILL — the commentary says a hero died / was taken down / got picked off.
{
  "type": "kill",
  "line": <int>,                 // the line where it is said
  "span": "<verbatim words>",
  "victim":      {"hero": "...", "player": "..."} | null,   // who died
  "killer":      {"hero": "...", "player": "..."} | null,   // who got the kill
  "killer_team": "<team code>" | null,   // use when credit goes to a TEAM and no player is
                                          // named: "first blood to AUR", "a kill for TEC"
  "outcome": "kill" | "survive"          // "survive" = the commentary says they got away
}
  · Name only what the commentary names. Leave a slot null rather than guessing — the
    checker scores "said the wrong person" very differently from "did not say".
  · Fill "killer" ONLY when the commentary CREDITS the kill to that person. Being nearby,
    or being the direction someone runs, is NOT credit.
  · Casters rarely say the word "kill". All of these mean someone died, record them:
    "he's dead" / "down goes X" / "X will die" / "X gets picked off" / "first blood, X" /
    "bursts him down" / "melts him" / "X finds one" / "that's a kill for X" / "X falls".
  · "flashes out" / "gets away on a sliver" -> outcome "survive".
  · Skip pure predictions ("if they collapse he dies"). Live play-by-play of something
    happening now is NOT a prediction: "he's going down", "X is dead" are kills.

OBJECTIVE — a team takes the Tyrant or the Overlord.
{
  "type": "objective",
  "line": <int>, "span": "<verbatim words>",
  "team": "<team code>" | null,
  "objective": "dragon" | "baron"
}
  · Vocabulary mapping — the checker uses the canonical LoL words:
      Tyrant / Dark Tyrant / Shadow Tyrant / 暴君  ->  "dragon"
      Overlord / Shadow Overlord / 主宰            ->  "baron"
  · ONLY these two. Turrets, inhibitors/crystals and the base are NOT annotated in the
    ground truth, so they cannot be checked — skip them entirely.
  · Only actual takes. Skip "they want the Overlord", "Tyrant spawns in 30 seconds".

Repeating the same kill across consecutive lines is ONE claim (use the first line).
Output JSON only, no markdown fences.
\end{Verbatim}
\end{CJK*}

\paragraph{Shared user-message template.}
\begin{Verbatim}
ROSTER (the only entities that exist in this game)
{roster}

COMMENTARY ({n} lines)
{lines}

Extract kill and objective claims. JSON only.
\end{Verbatim}

\endgroup

\section{Locator Quality}
\label{app:locator}
Table~\ref{tab:locator} compares anchors against automatic reference
annotations on SFT outputs. The specialized locator improves Exact
agreement from 47.9\% to 56.7\%; combining it with the scorer reaches
57.9\% without detection gating. On the reported 360-event diagnostic
subset, telemetry timestamps achieve only 6.4\% Exact agreement.
This gap supports locating event mentions in the commentary rather
than assuming they coincide with event timestamps
(Figure~\ref{fig:credit_ablation}).

\begin{table}[H]
\centering
\small
\setlength{\tabcolsep}{6pt}
\caption{Anchor agreement with automatic reference annotations on SFT
outputs. Exact and $\pm1$\,s agreement are percentages; MAE is in seconds.
Learned variants use 363 events without detection gating; the telemetry
row uses a 360-event diagnostic subset. A dash denotes an unreported
metric.}
\label{tab:locator}
\begin{tabular}{lrrrr}
\toprule
Method & Events & Exact$\uparrow$ & $\pm1$ s$\uparrow$ & MAE$\downarrow$ \\
\midrule
Telemetry timestamp & 360 & 6.4 & 20.3 & -- \\
Scorer as locator & 363 & 47.9 & 57.3 & 4.41 \\
Specialized locator & 363 & 56.7 & 66.4 & 3.34 \\
Scorer + locator & 363 & \textbf{57.9} & \textbf{67.5} & \textbf{3.20} \\
\bottomrule
\end{tabular}
\end{table}

\section{Game Agent Details}
\label{app:game-agent}

This appendix provides the experimental details and additional analyses
for the gameplay experiments in Section~\ref{sec:game-agent}.

\subsection{Environment and Data Collection}

We use Honor of Kings as the environment for both training and evaluation. To support
real-time evaluation and interaction, we develop a GUI-based framework in which model inference is
deployed on a remote server, while live matches are run through Tencent Mobile Game Assistant on a
Windows client, with low-latency communication between the two components. At each interaction
step, the Windows client captures a screenshot of the current game window and transmits it to the
remote server for inference. The server then sends the model-predicted action back to the Windows
client, which executes the action in the game.

To collect human demonstrations for Honor of Kings, we build a dedicated platform for
real-time gameplay data collection. Participants play the game through Tencent Mobile Game
Assistant on Windows using a keyboard and mouse. Every 50~ms, the system records a screenshot of
the game window together with the player's mouse actions. This process yields approximately
1,000 minutes of human gameplay data.

To expand the training dataset, we follow the data expansion strategy of Video PreTraining
(VPT)~\citep{baker2022vpt} and train an inverse dynamics model (IDM) based on Qwen3.5 using
the collected human demonstrations. Given a sequence of 32 gameplay frames, our IDM predicts the
actions corresponding to the central 16 frames. We additionally collect approximately 24 hours of
human gameplay livestream recordings from the Internet and use the trained IDM to generate action
labels for these videos. We then clean the combined dataset using visual recognition techniques
based on template matching, yielding more than 2,000 minutes of training data from directly
recorded human demonstrations and automatically labeled livestream videos. In all collected
gameplay data, the player-controlled hero is Daji.

\subsection{Agent Architecture}

The real-time nature of Honor of Kings imposes stringent latency requirements, whereas
autoregressive decoding with a VLM incurs substantial inference overhead. To address this challenge,
we design a low-latency asynchronous dual-system architecture that couples our MOBA-VL
with NVIDIA's NitroGen~\citep{magne2026nitrogen}, drawing inspiration from
OpenHelix~\citep{cui2025openhelix}.

Specifically, we append eight learnable tokens to the original input sequence of our
MOBA-VL model. A single forward pass yields the hidden states at these eight token positions,
which are used as high-dimensional latent guidance for action generation. The NitroGen component
comprises a vision encoder and a diffusion transformer (DiT)~\citep{DBLP:conf/iccv/PeeblesX23} trained with flow matching~\citep{DBLP:conf/iclr/LipmanCBNL23}. We
concatenate the VLM latent tokens with the visual tokens from NitroGen's vision encoder and feed
this combined conditioning sequence to NitroGen's DiT to generate the final actions.

At inference time, the NitroGen-based fast system processes the latest game image and predicts
actions at every interaction step. The MOBA-VL-based slow system receives a new game image
and updates its latent guidance once every six inference calls of the fast system. Between updates,
the fast system reuses the most recently available VLM latents while continuing to respond to new
visual observations. During inference, we use NitroGen's default action space, action horizon,
and flow-matching steps.

During training, we freeze the pretrained weights of the VLM and optimize other parameters.
This setup allows us to assess whether our previously trained VLM can support gameplay and
provide useful high-dimensional action guidance.

\subsection{Evaluation Protocols}

We evaluate our model using both open-loop and closed-loop protocols. For comparison, we train
two baselines using the same dataset and training hyperparameters as our model:
(i) the original NitroGen model without a VLM, and (ii) a dual-system variant in which the
MOBA-VL weights are replaced with Qwen3.5-9B weights.

For open-loop evaluation, we split the combined dataset of recorded human gameplay and Internet
gameplay videos into a train set and a test set. The models are trained on the train set and
evaluated on the test set. For each sample in the test set, we compute the mean squared error
(MSE) between the predicted actions and the corresponding ground-truth action labels.

\begin{figure}[htbp]
\centering
\includegraphics[width=\linewidth]{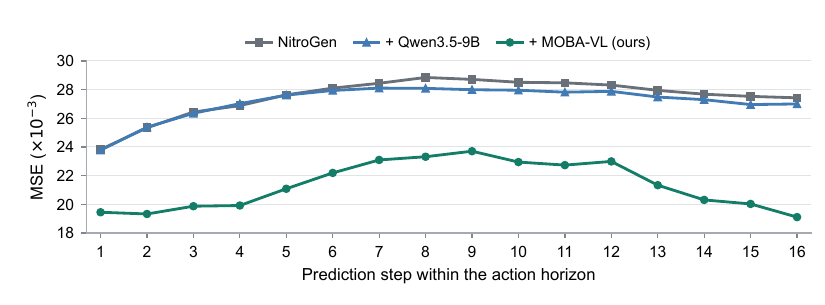}
\caption{Open-loop MSE at each prediction step within the 16-step action horizon.}
\label{fig:game-agent-open-loop-horizon}
\end{figure}

Figure~\ref{fig:game-agent-open-loop-horizon} shows lower error for MOBA-VL + NitroGen
at all 16 prediction steps, indicating that the gains extend across the action horizon.

For closed-loop evaluation, we additionally collect approximately 80 minutes of early-game
Honor of Kings gameplay videos in which players control the hero Daji, and use our IDM
to annotate these videos with action labels. Starting from the weights obtained in the open-loop
experiments, we further fine-tune each model on this dataset for three epochs using the same
training hyperparameters. We then deploy the resulting models on our real-time gameplay platform
to play matches against in-game bots. For each match, we record the first 130 seconds of gameplay
and use \texttt{GPT-6 Astra} via its API to score the recorded video, thereby quantifying the
performance of different models in closed-loop interaction.

The scoring rubric covers equipment purchase, skill upgrades, control smoothness, movement
strategy, attack timing, and teammate support.
The open-loop and closed-loop results are reported in
Tables~\ref{tab:game-agent-open-loop} and~\ref{tab:game-agent-closed-loop}, respectively.

\subsection{Latent Analysis}
\label{app:game-agent-latent}

\begin{figure}[H]
\centering
\includegraphics[width=\linewidth]{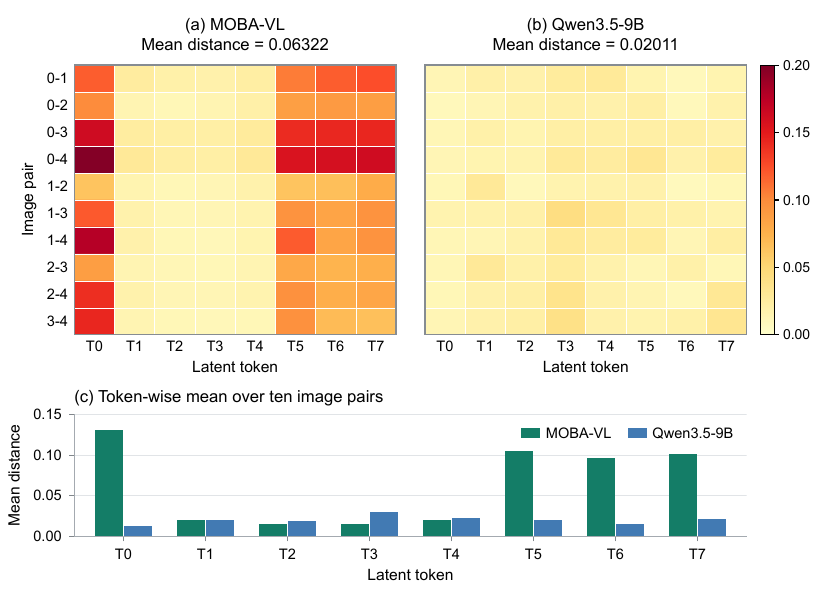}
\caption{Input dependence of VLM latent tokens for five test images. (a--b) Each cell shows
the cosine distance between the corresponding token vectors for one image pair, using a
shared color scale. (c) Mean distance at each token position across the ten image pairs.
Larger distances indicate greater variation across input images.}
\label{fig:game-agent-latent-analysis}
\end{figure}

To further examine whether the VLM produces informative latent tokens, we select five samples
from the test set and feed the corresponding images to the VLM to extract their latent tokens.
For each of the ten distinct image pairs, we compute cosine similarity between the corresponding
latent vectors at each of the eight token positions. We report cosine distance,
\[
d_{\cos}(\mathbf{u},\mathbf{v})
= 1 - \frac{\mathbf{u}^{\top}\mathbf{v}}{\|\mathbf{u}\|_2\,\|\mathbf{v}\|_2},
\]
and perform this analysis separately with
the MOBA-VL and Qwen3.5-9B backbones.
Figure~\ref{fig:game-agent-latent-analysis} reports the resulting distances,
including token-wise means averaged across the ten image pairs. MOBA-VL
has a mean cross-image cosine distance of 0.06322, versus 0.02011 for
Qwen3.5-9B, showing greater input-dependent variation.

\FloatBarrier

\end{document}